\documentclass[11pt]{article}

\usepackage{2_arXiv_Version/my_packages}
\usepackage{2_arXiv_Version/my_commands}
\usepackage{geometry}

\usepackage[numbers,sort&compress]{natbib}
\usepackage{xcolor}
\usepackage[
  colorlinks=true,
  linkcolor=blue,
  citecolor=teal,
  urlcolor=magenta
]{hyperref}

\graphicspath{{0_figures/}}
\title{Lightweight Tracking Control for Computationally Constrained Aerial Systems with the Newton-Raphson Method}

\author{
Evanns Morales-Cuadrado,
Luke Baird,
Yorai Wardi,
and Samuel Coogan%
\thanks{The NASA University Leadership Initiative (grant \#80NSSC20M0161) provided funds to assist the authors with their research, but this article solely reflects the opinions and conclusions of its authors and not any NASA entity.}%
\thanks{The authors are with the School of Electrical and Computer Engineering at the Georgia Institute of Technology, \texttt{\{egm,lbaird38,ywardi,sam.coogan\}@gatech.edu}. S. Coogan is also with the School of Civil and Environmental Engineering.}
}

\date{}
\begin{document}
\maketitle

\begin{abstract}
We investigate the performance of a lightweight tracking controller, based on a flow version of the Newton-Raphson method, applied to a miniature blimp and a mid-size quadrotor. This tracking technique admits theoretical performance guarantees for certain classes of systems and has been successfully applied in simulation studies and on mobile robots with simplified motion models.
We evaluate the technique through real-world flight experiments on aerial hardware platforms subject to realistic deployment and onboard computational constraints. The technique's performance is assessed in comparison with established baseline control frameworks of feedback linearization for the blimp, and nonlinear model predictive control for both the quadrotor and the blimp. The performance metrics under consideration are (i) root mean square error of flight trajectories with respect to target trajectories, (ii) algorithms' computation times, and (iii) CPU energy consumption associated with the control algorithms. The experimental findings show that the Newton-Raphson-based tracking controller achieves competitive or superior tracking performance to the baseline methods with substantially reduced computation time and energy expenditure.
\end{abstract}

\textbf{Keywords:} Tracking control, quadrotor applications, blimp applications, model predictive control, feedback linearization.

\section{Introduction}\label{sec:intro}
The past two decades have witnessed a significant shift in the nature of hardware research for trajectory control of aerial platforms like quadrotors. Initially, testing and verification of novel techniques relied heavily on numerical simulators, later transitioning to real-world deployments that depended on ground station computers and simplified models; see~\cite{realtimeMPC_2014}. Later developments of powerful single-board computers have enabled research to shift towards onboard execution even for computationally intensive control techniques~\cite{nmpc_so3, nmpc_faulttolerant_scaramuzza, scaramuzza_nmpc_dfbc}. The use of such computers can be credited for the maturation of several control techniques, of which Nonlinear Model Predictive Control (NMPC) has emerged as the state-of-the-art framework for quadrotor control~\cite{scaramuzza_l1_nmpc}.

Advances in computing have also impacted the development of controllers for other kinds of Unmanned Aerial Vehicles (UAVs).
In the realm of blimp research, older methods~\cite{HF-KK-YH-FM-KK-HA:2007} were computationally limited to simplified linear models in order to perform model predictive control, but more recent works have reported on implementations of NMPC with high-fidelity models~\cite{MK-LB-SC:2024}. Reinforcement learning-based controllers that improve upon existing PID tracking controllers have also been implemented for blimps~\cite{TL-EP-MB-AA:2022}.

However, much of the state-of-the-art research has become dependent on exploiting powerful but relatively costly computing platforms. This renders them unrealistic for certain applications.
As noted in~\cite{scaramuzza_nmpc_dfbc}, it may be ``impractical to run NMPC on some miniature aerial vehicles with a limited computational budget.''

All of this suggests an opening for exploration of effective tracking-control techniques that are computationally lightweight enough so as not to require top-of-the-line hardware for their onboard implementations. In particular, such techniques ought to be sufficiently general to be easily deployable on various hardware platforms with minimal alterations, and be capable of low-level control at high frequencies.

\begin{figure}[t]
    \centering
    \includegraphics[width=0.50\textwidth]{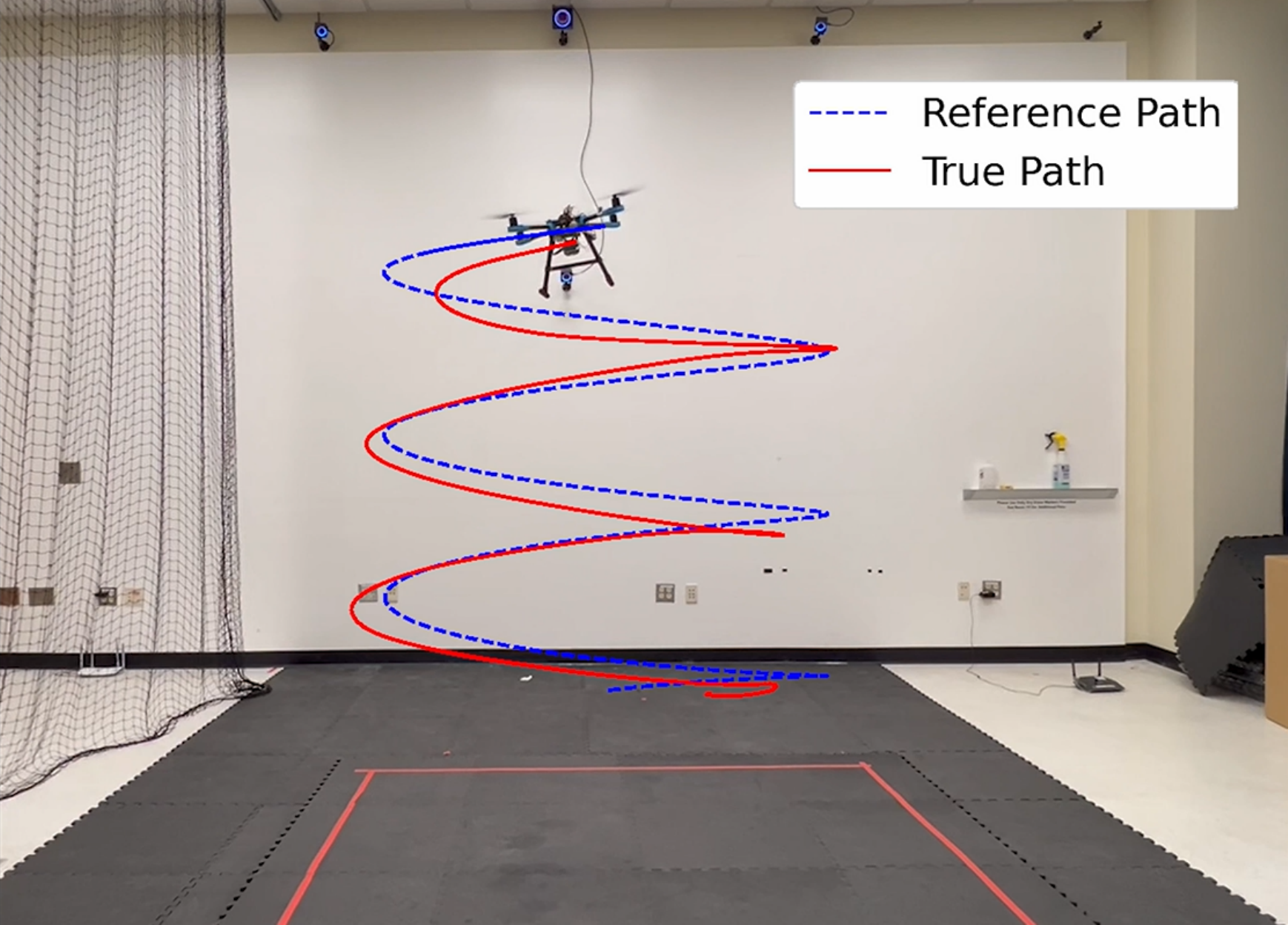}
    \caption{Helix trajectory tracking with the Newton-Raphson-based tracking technique.}
    \label{fig:intro_image}
\end{figure}

With this in mind, authors of this paper have investigated a recently-proposed concept for tracking control that is based on efficient computations.
It is comprised of three elements: the Newton-Raphson method for computing the roots of equations, a single-instance lookahead for predictions, and a simple mechanism for speeding up the resulting controller's action. All three elements will be explained in detail in the sequel, where we argue for the efficiency and effectiveness of their combination as defined in \cite{8264633, wardi_ijrnc_2019}, for suitable applications.

The main objective of this paper is to investigate comparative performance tradeoffs between the tracking technique proposed in \cite{wardi_ijrnc_2019} and established tracking controllers which have had widespread success in applications: NMPC for quadrotors and blimps, and additionally, feedback linearization for blimps. The considered comparative performance metrics are Root Mean Square (tracking) Errors (RMSE), CPU times required by the controllers, and the energy spent by them during common control experiments.
Results of several experiments have been tabulated and explained in the sequel, and they
suggest that the tracking technique presented in
\cite{wardi_ijrnc_2019} may be a viable alternative to the aforementioned established techniques under certain conditions. An illustration of the technique in action is provided in Fig.~\ref{fig:intro_image}, which depicts quadrotor trajectory tracking using the proposed method.

The rest of the paper is organized as follows. Section \ref{sec:methodology} explains the tracking methodology proposed in \cite{wardi_ijrnc_2019}, recounts relevant results from the literature, provides a statement of innovation in the present paper, and comments on implementation details for
its experimental framework. Section \ref{sec:experiments_blimp} and Section \ref{sec:experiments_quadrotor} present experimental results for a blimp and a quadrotor, respectively, and Section \ref{sec:conclusion} summarizes the results and concludes the paper.\footnote{Our code and video demonstrations can be found at: \url{https://github.com/gtfactslab/MoralesCuadrado_TCST2025}}

\section{Methodology}\label{sec:methodology}
The tracking-control technique presented in this paper was first introduced in \cite{8264633} and further developed in \cite{wardi_ijrnc_2019}. Much of the initial discussion in this section follows \cite{wardi_ijrnc_2019}, and consequent results are summarized in later citations.

We refer to the tracking control technique proposed in \cite{wardi_ijrnc_2019} by the ``NR-based tracking technique'', or ``NR-based technique'' for brevity. The term ``NR'' highlights the fact that the tracking controller is founded on the Newton-Raphson method for finding roots of equations, as described next.

\subsection{Fundamental ideas underscoring the NR-based tracking technique}\label{subsec:methodology_basic_ideas}
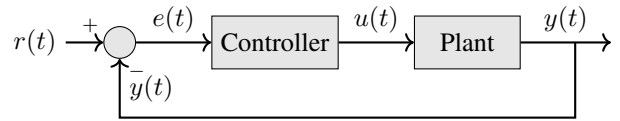
\begin{figure}[h]
    \centering
    \begin{tikzpicture}[
      block/.style = {draw, fill=gray!20, rectangle, minimum height=2em, minimum width=4em},
      sum/.style   = {draw, circle, fill=gray!20, minimum size=1.2em, inner sep=0pt},
      arrow/.style = {->, thick},
      node distance=.4cm and 1cm,
    ]

    \node[sum] (sum) {};
    \node[block, right=of sum] (controller) {Controller};
    \node[block, right=of controller] (plant) {Plant};
    \node[right=1.2cm of plant] (state) {};
    \node[left=1.2cm of sum] (ref) {};

    \draw[arrow] (ref) -- (sum)
        node[midway, above, yshift=1pt] {\scriptsize $r(t)$};

    \draw[arrow] (sum) -- (controller)
        node[midway, above] {\scriptsize $e(t)$};

    \draw[arrow] (controller) -- (plant)
        node[midway, above] {\scriptsize $u(t)$};

    \draw[arrow]
      (plant) --
      node[midway, above] {\scriptsize $y(t)$}
      coordinate[midway] (ytmid)
      (state);

    \coordinate (fbottomright) at ($(ytmid)+(0,-1.3cm)$);
    \coordinate (fbottomleft)  at ($(sum)+(0,-1.3cm)$);

    \draw[arrow]
      (ytmid) -- (fbottomright)
      -- node[midway, above] {\scriptsize $y(t)$}
      (fbottomleft)
      -- (sum);

    \node at ($(sum)+(-.45,-0.15)$) {\scriptsize $+$};
    \node at ($(sum)+(-0.22,-.4)$) {\scriptsize $-$};

    \end{tikzpicture}
    \caption{{\small Basic feedback control system.}}
    \label{fig:basic_control_sys}
\end{figure}

The setting for the tracking technique is a continuous-time dynamical system defined by an ordinary differential equation as depicted in Figure \ref{fig:basic_control_sys}, where $u(\cdot)$, $y(\cdot)$ and $r(\cdot)$ denote the input to the plant, output of the system, and reference signal, respectively. We assume that $t\in[0,\infty)$, and that for every $t\geq 0$, $u(t),~y(t)$ and $r(t)$ are in a Euclidean space $\mathbb{R}^{m}$ of a common dimension $m>0$. The state variable of the plant, $x(t)\in\mathbb{R}^n$, is internal to the plant, and its dimension, $n>0$, need not be equal to $m$. The objective of the controller is to ensure that the output $y(\cdot)$ tracks the reference signal $r(\cdot)$ in a suitable sense defined below.

We start the discussion with the following simple scenario in order to highlight the basic ideas behind the NR-based tracking technique. Suppose that the plant subsystem in Figure \ref{fig:basic_control_sys} is defined by a memoryless nonlinearity of the form
\begin{equation}
\label{eqn:memoryless_plant}
    y(t)=g(u(t)),
\end{equation}
where the function $g:\mathbb{R}^{m}\rightarrow \mathbb{R}^{m}$ is continuously differentiable.
Suppose also that the target trajectory is a constant, namely $r(t)\equiv r$ for a given $r\in\mathbb{R}^m$. Then the tracking controller is defined as
\begin{equation}
\label{eqn:simple_NRflow}
    \dot{u}(t)=\left(\frac{dg}{du}(u(t))\right)^{-1}\bigl(r-g(u(t))\bigr),
\end{equation}
and we recognize it as a continuous-time flow version of the traditional iterative Newton-Raphson method for solving the equation $r-g(u)=0$ in the variable $u\in\mathbb{R}^m$,
\[   u_{i+1} = u_i - \left(\frac{\partial g}{\partial u}(u_i)\right)^{-1}\Bigl(r_i - g(u_i)\Bigr),
\]
$i=0,1,2,\ldots$.
For further discussion on the relationship between iterative algorithms and their ``fluid-flow'' variants, please see \cite{wardi_ijrnc_2019}.

In the present discussion we implicitly assume that the Jacobian $\frac{dg}{du}(u)$ is nonsingular along the trajectory $u(\cdot)$ and other regularity conditions are satisfied, as pointed out in \cite{wardi_ijrnc_2019}.

Define the function $V:\mathbb{R}^m\rightarrow\mathbb{R}^+$
by
\begin{equation}
    V(u):=\frac{1}{2}||r-g(u)||^2.
    \label{eqn:tracking_lyapunov}
\end{equation}
Under the regularity assumptions in~\cite{wardi_ijrnc_2019}, convergence of $u(t)$ to a root of the equation $r-g(u)=0$ is proven by taking the derivative $\dot{V}(u(t))$ in (\ref{eqn:tracking_lyapunov}).
To see this point, combine~\eqref{eqn:simple_NRflow}~--~\eqref{eqn:tracking_lyapunov} to obtain
\begin{align}
\label{eqn:Vdot_tracking}
    \dot{V}(u(t)) \nonumber  &=-\big\langle r-g(u(t)),\frac{dg}{du}(u(t))\dot{u}(t)\big\rangle \nonumber  \\
    &=-\big\langle  r-g(u(t)),r-g(u(t)) \big\rangle \nonumber \\
    &=-||r-g(u(t))||^2=-2V(u(t)),
\end{align}
where the function $\langle\cdot,\cdot\rangle$ stands for the standard
inner product in $\mathbb{R}^m$.
This establishes the fact that $V(u(t))$ is a Lyapunov function,
and a convergence of the output to the (constant) reference has the form
\begin{equation}
\label{eqn:asymptotic_tracking}
    \lim_{t\rightarrow\infty}y(t)=r.
\end{equation}

Next, we relax the assumption that $r(\cdot)$ is a constant
and assume instead that $r(\cdot)$ is a bounded, continuous, and piecewise-continuously differentiable function of time $t$ with bounded derivative $\dot{r}(t)$ over $t\in[0,\infty)$. We revise the definitions in \eqref{eqn:simple_NRflow} and~\eqref{eqn:tracking_lyapunov} by replacing $r$ by $r(t)$, but the corresponding results in~\eqref{eqn:Vdot_tracking} and~\eqref{eqn:asymptotic_tracking} are no longer guaranteed. However, the following weaker conclusion is in force:
\begin{equation}
\label{eqn:r_t_tracking}
    \limsup_{t\rightarrow\infty}||r(t)-y(t)||~\leq~\limsup_{t\rightarrow\infty}||\dot{r}(t)||.
\end{equation}
Observe that~\eqref{eqn:asymptotic_tracking} is the special case of~\eqref{eqn:r_t_tracking} where $r(\cdot)\equiv r$, a constant.

A tightening of the Right-Hand Side (RHS) in~\eqref{eqn:r_t_tracking}
can be obtained by scaling the RHS of the control equation~\eqref{eqn:simple_NRflow} by a constant $\alpha>1$.

With this scaling the controller has the following form,
\begin{equation}
\label{eqn:speedup_nr_flow}
    \dot{u}(t)=\alpha\left(\frac{dg}{du}(u(t))\right)^{-1}\big(r(t)-g(u(t))\big),
\end{equation}
and consequently, the conclusion expressed by~\eqref{eqn:r_t_tracking} is generalized to the inequality
\begin{equation}
\label{eqn:r_t_tracking_with_speedup}
 \limsup_{t\rightarrow\infty}||r(t)-y(t)||~\leq~\alpha^{-1}\limsup_{t\rightarrow\infty}||\dot{r}(t)||.
\end{equation}
It is noted in~\cite{wardi_ijrnc_2019} that the constant $\alpha$ acts as a speedup factor, not a gain of the controller subsystem, since it is a multiplicative factor of $\dot{u}(t)$, not $u(t)$.
\vspace{.1in}

The final stage of the discussion in this subsection considers the most general case of the plant subsystem, modeled by a differential equation of the form
\begin{equation}
\label{eqn:general_dynamical_system}
    \dot{x}(t)=f(x(t),u(t)),~~~~~~x(0)=x_{0},
\end{equation}
and an output function
\begin{equation}
    y(t)=h(x(t)).
    \label{eqn:output_function}
\end{equation}
In the state equation~\eqref{eqn:general_dynamical_system}, $x(t)\in\mathbb{R}^n$ is the state variable
and $u(t)\in\mathbb{R}^m$ is the plant's control input. The function $f:\mathbb{R}^n\times\mathbb{R}^m\rightarrow\mathbb{R}^n$ is assumed to have sufficient smoothness and boundedness properties for the existence of unique solutions of the differential equation~\eqref{eqn:general_dynamical_system} for all $t\in[0,\infty)$, for every $u(\cdot)$ and $x_{0}$ in respective sets of interest (see \cite{wardi_ijrnc_2019} for details). The system's output function $h:\mathbb{R}^{n}\rightarrow\mathbb{R}^m$ is assumed to be
continuous and piecewise-continuously differentiable with bounded first derivatives on compact sets in $\mathbb{R}^n$.

Regarding the effects of this dynamic formulation on the NR-based tracking controller, the main difference between the memoryless system~\eqref{eqn:memoryless_plant} and the dynamic model~\eqref{eqn:general_dynamical_system}~--~\eqref{eqn:output_function} is that in the latter case
$y(t)$ is not a function of $u(t)$, therefore it is impossible to use~\eqref{eqn:speedup_nr_flow} to define the controller. However, for given $t\geq 0$ and $T>0$, both $x(t+T)$ and $y(t+T)$ are functions of $x(t)$ and $\{u(\tau):\tau\in[t,t+T)\}$. Therefore it is reasonable to define, at time $t$,
a predictor of $y(t+T)$ with the objective of having it track $r(t+T)$.

Since the future inputs $\{u(\tau):\tau\in(t,t+T]\}$ are unknown at time $t$, we approximate them by assuming that the control input is held constant over the prediction horizon, i.e., a zero-order hold approximation is used. Under this assumption we consider a predicted output, denoted by
$\tilde{y}(t+T)$, as a function of $x(t)$ and $u(t)$. That is, we have that
\begin{equation}
\label{predictor}
    \tilde{y}(t+T)=g(x(t),u(t))
\end{equation}
for a continuous function
$g:\mathbb{R}^n\times\mathbb{R}^{m}\rightarrow\mathbb{R}^m$,
assumed to be continuously differentiable
in $(x,u)$ and whose partial Jacobian
$\frac{\partial g}{\partial u}(x(t),u(t))$ is
locally Lipschitz continuous in $(x,u)$. Note that for this Jacobian to be invertible and for tracking error to be computed, the reference, system inputs, and outputs must all be of the same dimension.

Given a fixed controller-speedup factor
$\alpha\geq 1$ and prediction horizon $T>0$,
the NR-based tracking control
technique is defined by the
following extension of
(\ref{eqn:speedup_nr_flow}),
\begin{equation}
\label{eqn:dynamic_system_NRflow_withspeedup}
    \dot{u}(t)=\alpha\left(\frac{\partial g}{\partial u}(x(t),u(t))\right)^{-1}\bigl(r(t+T)-g(x(t),u(t))\bigr).
\end{equation}
The objective of this controller is to enable $\tilde{y}(t+T)$ to track the future input $r(t+T)$\footnote{The future reference $r(t+T)$ is defined as the tracking-target at time $t+T$, regardless of how it is computed. For instance, in a simple case $r(\cdot)$ may be an exogenous process that is explicitly known to the system at the initial time $t_{0}:=0$. In a more involved scenario, $r(t+T)$ may be a predicted value of an endogenous process of the system.}.
The effectiveness of the NR-based tracking technique is characterized by the following inequality,
\begin{equation}
\label{eqn:r_tT_tracking_with_speedup}
\begin{aligned}
 \limsup_{t \rightarrow \infty}\|r(t+T)-\tilde{y}(t+T)||&\\
 \leq\alpha^{-1} \limsup_{t \rightarrow \infty} \|\dot{r}(t)\|,&
\end{aligned}
\end{equation}
where we note the role of $\alpha$ in scaling down the RHS of (\ref{eqn:r_tT_tracking_with_speedup}), which is independent of $T>0$.

By combining~\eqref{eqn:general_dynamical_system} and~\eqref{eqn:dynamic_system_NRflow_withspeedup}, the closed-loop system can be viewed as a dynamical system with the compounded state variable $z:=(x^{\top},u^{\top})^{\top}\in\mathbb{R}^{n}\times \mathbb{R}^m$ and input $r(\cdot)$. Unlike the situation with the memoryless plant system defined by~\eqref{eqn:memoryless_plant}, Bounded Input Bounded State (BIBS) stability cannot be taken for granted or proved from verifiable assumptions.
Furthermore, \cite{wardi_ijrnc_2019} defines a uniform variant of BIBS stability, labeled {\it $\alpha$ stability}, and proves from it the following limit,
\begin{equation}
\label{eqn:uniform_asymptotic_tracking}
\lim_{\alpha\rightarrow\infty}\limsup_{t\rightarrow\infty}||r(t)-\tilde{y}(t)||~=~0.
\end{equation}
\eqref{eqn:uniform_asymptotic_tracking} provides a definition for uniform asymptotic tracking of the NR-based technique.
The theoretical question of tracking now hinges on whether $\alpha$ stability can be proven from verifiable assumptions for a particular system or a class of systems. That has been achieved for linear, time-invariant systems in \cite{wardi_ijrnc_2019}, but proofs for nonlinear systems remained elusive until recently, as discussed in the sequel.

\subsection{Additional Results}

While the theoretical question of $\alpha$ stability has been slow to yield provable results, investigations of the NR-based technique focused primarily on simulation studies of various systems. Of a particular interest are dynamic models of autonomous vehicles,
arrayed in platoons and other formations, with distributed motion controls by the NR-based tracking technique.

Simulation results of experiments on platoons of the dynamic bicycle model
\cite{falcone_tcst_2007_active_steering} are contained in~\cite{shivam_ecc2019_autonomous_vehicles,shivam_ifac2020_intersection_traffic}. \cite{niu_lcss_2023_consensus_output_prediction} extended these ideas to heterogeneous multi-agent networks with graph-Laplacian--based motion control, while \cite{NN_Pred_1} studied a model-free setting in which agent dynamics are learned directly from data.
Over the course of these simulation studies, various ideas have been proposed to enhance the core NR-based tracking method. These include the incorporation of differential flatness
for some vehicles~\cite{notomista_acc2024_safe_tracking}, the use of machine learning methods to improve model accuracy, and the development of a barrier function approach for dynamic control laws~\cite{ICBF}.

In these simulation experiments
we make the following observations.
\begin{enumerate}
    \item
At a given $\alpha>0$, systems became unstable when the prediction horizon $T$ was taken to be too short.
\item
At a given $T>0$, increasing $\alpha$ serves to stabilize an otherwise unstable system.
\item
Given $T>0$ and $\alpha>0$ such that the closed-loop system is stable, if a system's predicted output is far off the target trajectory at the initial time, namely $||\tilde{y}(T)-r(T)||$ is large, then the closed-loop system may exhibit substantial overshoots and oscillations in the input $u(t)$, state variable $x(t)$, and/or output $y(t)$.
\end{enumerate}

The first and second observations
led us to the following process for choosing $T$ and $\alpha$ (see \cite{wardi_ijrnc_2019}):
\begin{itemize}
\item
{\it Choose a small $T>0$ so as to achieve a small asymptotic tracking error if the closed-loop system were stable. If the closed-loop system is unstable, increase $\alpha$ to stabilize it.}
\end{itemize}
This approach of choosing $T$ and $\alpha$ has been used in most of the aforementioned simulation experiments and shown to be effective.

The third observation above led to the use of a particular version of Control Barrier Functions (CBF), suitable for dynamic controllers~\cite{ICBF}. Labeled Integral CBF, or I-CBF, it extends the scope of CBF to include systems whose plant's dynamics are not control affine, as for the dynamic bicycle.\footnote{A dynamic control directly computes $\dot{u}(t)$, not $u(t)$. This requires an integrator to generate the input to the plant, $u(t)$, hence the \emph{I-CBF} label.} Furthermore, its application to dynamic controllers often requires lighter computing loads than the standard CBF.

A notable challenge in tracking control arises in the context of nonholonomic systems, i.e., systems subject to velocity constraints that cannot be integrated into positional constraints. Such systems are common in robotics, for instance wheeled vehicles whose motion is restricted to their instantaneous heading direction. Additionally, for many of these systems the dimension of the input is strictly smaller than that of the output of interest, which violates the equal-dimension requirement of the NR-based technique. A specific instance of this difficulty is addressed for the NR-based technique in~\cite{shivam_cdc2019_pursuit_evasion}, where Dubins-car pursuers in a pursuit--evasion game are controlled by collapsing the two-dimensional position output into a scalar distance metric, thereby resolving the dimensional mismatch in that particular setting. However, extending the NR-based technique to a broader class of nonholonomic systems remains an open direction for future research. Notably, recent developments have extended NMPC to hardware applications on nonholonomic systems~\cite{MR_2023}.

An approach to the NR-based tracking controller by the concept and technique of differential flatness is also investigated in~\cite{notomista_acc2024_safe_tracking}. Under broad verifiable assumptions, a quasi NR-based controller, operating on the flat system, can perform the prediction, I-CBF, and tracking control with greatly reduced computational burden as compared to their implementations on the physical system. In contrast,~\cite{niu_cdc2025_stability_flat_arxiv} uses the differential flatness only for the prediction but not the I-CBF or NR-based controller. Thus, it preserves the NR-based technique and not approximations thereof, but it requires more computing efforts than the aforementioned quasi NR-based technique. While proofs of \(\alpha\)-stability are established in~\cite{wardi_ijrnc_2019} for linear systems,~\cite{niu_cdc2025_stability_flat_arxiv} provides a proof of $\alpha$-stability for differentially flat systems for the case where the target trajectory $r(\cdot)$ is a constant $r\in\mathbb{R}^m$.

The earliest implementation of the NR-based tracking technique on quadrotors was reported in \cite{mypaper}, where the quadrotors are the same as those considered in this paper. The experiments in \cite{mypaper} tested a simple output prediction method, comprised of a linear, time-invariant prediction model, and used it to test the effectiveness of the tracking technique. The resultant tracking was not as precise as in this paper (see Subsection II.D, below), which uses a more sophisticated prediction model, and it did not include tests on other aerial vehicles. Also, \cite{mypaper} does not perform any comparisons with powerful, established tracking control alternatives, which is the main objective of this paper.

\subsection{Contributions of this paper}
Earlier in this section we mentioned a number of system models which have been simulated for the purpose of testing the NR-based tracking technique. At this point, we believe that the main challenge in the investigation of the NR-based tracking technique is to conduct a comparative study on hardware platforms of its performance and that of well-known tracking controllers that are commonly used in research and applications. Specifically, we are interested in comparative assessments when target trajectories require large computing loads. To this end, we perform the comparisons against variants of feedback linearization-based control and nonlinear model predictive control. The resulting comparisons were conducted in laboratory settings and the results are presented in the sequel.

\subsection{Implementation Details}

We close this section with a statement about the predictor used in most of the works on the NR-based technique.
Our previous work on quadrotors \cite{mypaper} focused on efficiency of computation to meet real-time requirements. To this end, it implemented the output prediction step of the control algorithm with a linearized closed-form solution to the system dynamics. By pre-computing the linearized system matrices at Euler angles of zero and assuming small deviations in applications, we saved time in the loop. This led to inaccuracies in output prediction and hence in tracking performance, as well as preventing us from considering more aggressive trajectories such as those containing yawing components.

In this work, we focus on achieving competitive tracking performance with powerful, established techniques on more aggressive and complex trajectories. Thus, we implement Runge-Kutta (RK4) and the forward Euler integration methods of the dynamics associated with the state equations of the blimp and quadrotor, respectively defined by~\eqref{eqn:blimp_full_dynamics_eqn} and~\eqref{eqn:quad_nonlindyn}, below.
The challenge in implementing such precise prediction methods is in ensuring that a complete iteration of the control algorithm is computed with a minimum update rate of \qty{100}{Hz}, even under the computational limitations of a Raspberry Pi.

To accomplish this, we employed two equally effective computational approaches using established software tools: (i) compilation of the integration and Jacobian computation routines into C via the Cython~\cite{behnel2011cython} toolchain and use as a Python shared object, and (ii) JIT compilation using the JAX~\cite{jax2018github} numerical computing library. In the Cython-based approach, Jacobians are computed by finite differences, whereas the JAX-based approach uses JIT-compiled forward-mode automatic differentiation. All experimental results showcased in this work use the JAX-based implementation for consistency.

This implementation allows for fast prediction with high accuracy, permitting us to compare results of the NR-based technique to those obtained from
NMPC and feedback linearization. Importantly, it also enables us to perform rotations in yaw decoupled from the motion of the quadrotor. This is key for applications that require the quadrotor to face a particular direction while it follows a path.

\begin{figure}[h]
    \centering
    \includegraphics[width=0.57\textwidth]{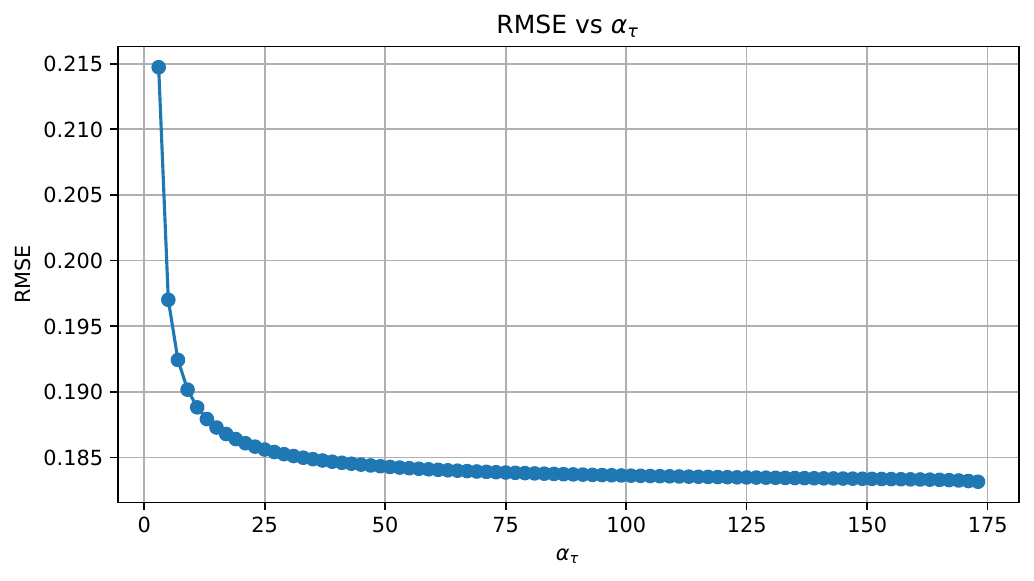}
    \caption{Empirical sensitivity of tracking error to the controller-speedup factor $\alpha$ associated with the thrust input $u_{\tau}$.}
    \label{fig:alpha_sensitivity}
\end{figure}

In addition, to assess the sensitivity of the NR-based technique to the speedup factor $\alpha$, we conduct a sweep over a wide range of values while holding all other parameters fixed. As shown in Fig.~\ref{fig:alpha_sensitivity}, increasing $\alpha$ beyond a moderate level yields only marginal improvements in tracking accuracy. Over a broad interval, the RMSE remains nearly constant, indicating that the controller performance is largely insensitive to the precise choice of $\alpha$ within this region. Outside the range of $\alpha$ depicted in the figure---both for sufficiently small and sufficiently large values---the closed-loop system becomes unstable. These tests were carried out in a numerical simulator of the quadrotor model utilized in this work, with standard ratios between the thrust speedup parameter and the remaining input parameters.

\section{Miniature Blimp}\label{sec:experiments_blimp}
Miniature blimps are an aerial platform that pose interesting challenges for control.
Blimps are constructed with a wide variety of actuation capabilities and envelope shapes, leading to a wide array of dynamic models.
Unlike quadrotors, blimps' neutral buoyancy allows them to stay aloft without any actuation, and thus they can be flown at a far cheaper energy cost for long-term missions compared to quadrotors.
Moreover, the dynamics of a blimp and a quadrotor are fundamentally different. Unlike the quadrotor, the blimp model considered in this paper has lightly damped zero dynamics in the rolling and pitching motions when input-output feedback linearized with positional outputs.

\begin{figure}[h]
    \centering
    \includegraphics[width=0.67\linewidth,trim={1cm .8cm 1cm 1.2cm}, clip]{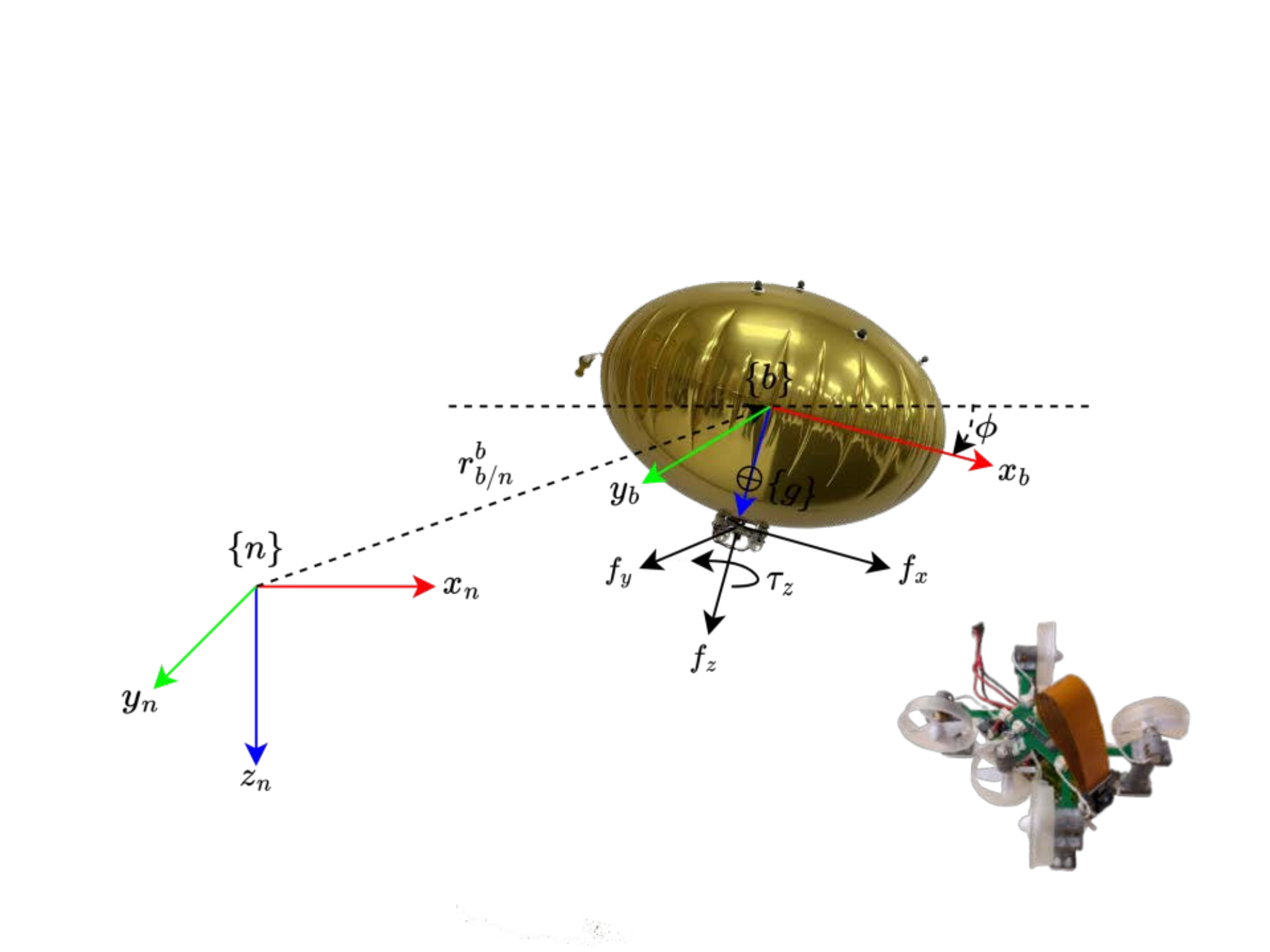}
    \caption{The radially symmetric blimp with relevant coordinate frames. The gondola includes six motors---two for vertical actuation, and four for horizontal actuation and yawing.}
    \label{fig:miniature_blimp_with_frames}
\end{figure}

Several tracking control strategies have been proposed for miniature blimps. The paper~\cite{HF-KK-YH-FM-KK-HA:2007} models a blimp as a simple chain of double integrators with time delay and tracks a desired yaw angle with model predictive control, but it is limited to only yaw tracking and ignores nonlinear dynamic effects.
Sliding mode control has been demonstrated to achieve good tracking performance~\cite{MW-AA-MC-IS-FS:2024}, but it requires the selection of the correct manifold onto which to project the dynamics, and it is computationally expensive.

In this work, we specifically consider a radially symmetric blimp with undermounted thrusters for horizontal holonomic control~\cite{QT-MH-FZ:2020}, for which several tracking controllers have been previously proposed. The paper~\cite{QT-JT-JC-YY-FZ:2021} presents a controller that tracks a desired horizontal velocity by tracking a fixed roll or pitch angle first in a multi-loop control architecture; however, this method is specifically useful for velocity tracking.
The paper~\cite{MK-LB-SC:2024} inverts the dynamics to derive a
feedback linearization (FBL)-based tracking controller. By including high-order control barrier functions (CBFs) to limit the roll and pitch angle---the residual zero dynamics---large roll and pitch angles are empirically mitigated.
The blimp and relevant coordinate frames are pictured in Figure~\ref{fig:miniature_blimp_with_frames}.

\subsection{Dynamic Model and Tracking Controllers for the Blimp}\label{subsec:blimp_model}

The radially symmetric blimp presented in~\cite{QT-MH-FZ:2020} is modeled as a six-DOF underactuated system with holonomic horizontal control. We briefly summarize the model---a more complete representation is given in~\cite{QT-JW-ZX-TL:2021,MK-LB-SC:2024}. Let \(\nu_{b/n}^b = (v_{b/n}^b,  \omega_{b/n}^b)\) represent translational and angular velocity in the body frame, and let \( \eta_{b/n}^n = (p_{b/n}^n, \Theta_{b/n}^n)\) represent position in the world frame and orientation of the blimp (expressed in Euler angles) relative to the world frame,
respectively. We then define the state vector for the blimp as $x=(\nu_{b/n}^b, \eta_{b/n}^n) = (v_{b/n}^b,  \omega_{b/n}^b , p_{b/n}^n, \Theta_{b/n}^n)\in\mathbb{R}^{12}$.
The inputs are forces along each axis in the body frame and torque about the body frame $z$-axis, $u = [f_x\,f_y\,f_z\,\tau_z]^\top$.

The dynamics about the center of gravity (CG) are originally derived from a submarine physics model~\cite{TF:2011} and are given here as

\begin{equation}
\label{eqn:blimp_full_dynamics_eqn}
\begin{split}
    \dot{x} =& \begin{bmatrix}
        -(M^{CB})^{-1} C^{CB}(x) - (M^{CB})^{-1} D^{CB} & 0\\
        \text{diag}(R(\Theta), T(\Theta)) & 0
    \end{bmatrix} \begin{bmatrix}
        \nu\\
        \eta
    \end{bmatrix} \\
    &- (M^{CB})^{-1} G^{CB}(x) + (M^{CB})^{-1} u
    \\
    y =& Wx
    \end{split}
\end{equation}
where \(M^{CB}\) is the mass/inertia matrix, \(C^{CB}(x)\) is the Coriolis-centripetal matrix, $D^{CB}$ is the diagonal aerodynamic damping matrix, $R(\Theta)$ is a rotation matrix, $T(\Theta)$ is a transformation matrix, $W$ is a binary matrix to select the position variables $p_x, p_y, p_z, \psi$,
and the gravity vector is
\begin{align*}
    G^{CB}(x) =
    -\begin{bmatrix}
        0_{3\times 1} \\
        r_{g/b}^b \times f_{g/b}^b
    \end{bmatrix}, \ f_g^b &= (R_b^n(x))^{-1}
    \begin{bmatrix}
        0 \\
        0 \\
        f_{z,g}^n
    \end{bmatrix}
\end{align*}
where \(f_{z,g}^n\) is the constant downward force of gravity and \(r_{g/b}^b = [0\ 0\ r^b_{z,g/b}]^\top\) is the position vector of the CG in the body frame, \emph{i.e.} the vector from the center of buoyancy (CB) to the CG. Although translational acceleration due to gravity is zero due to the blimp's buoyancy, gravity induces non-zero restoring torque about the CB.

In the case of the miniature blimp, we compare our proposed NR-based tracking technique against the FBL-based and NMPC control strategies developed in~\cite{MK-LB-SC:2024}.
To maintain consistency with the experimental setup in~\cite{MK-LB-SC:2024} and to enable a fair comparison with previously published results, we utilize the same hardware, motion-capture software, and implementation of these miniature blimp controllers. Our objective in this work is to evaluate the NR-based technique against established baseline controllers under their standard configurations as reported in the literature.

\begin{figure*}[t]
    \centering
    \includegraphics[width=0.95\textwidth]{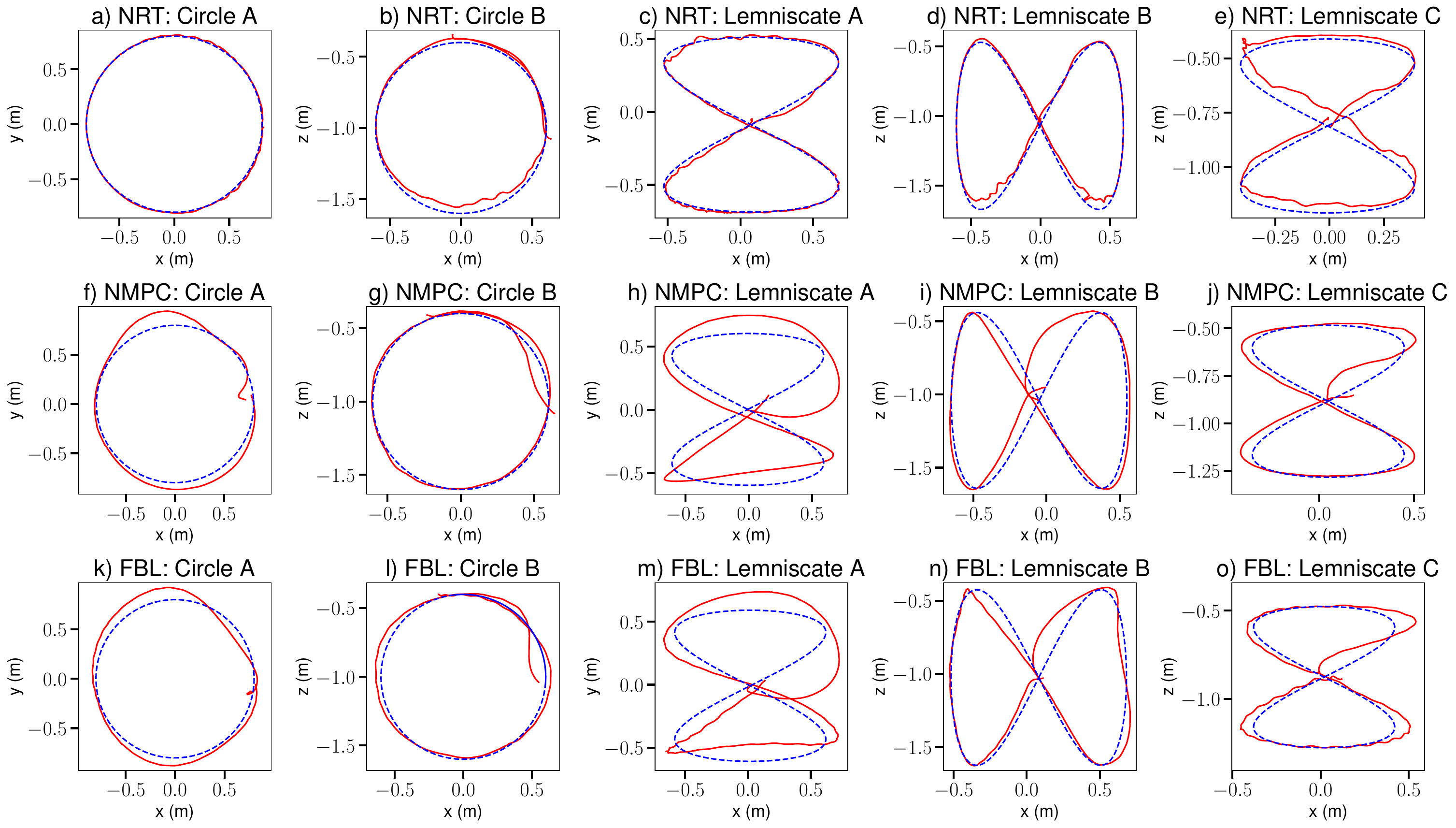}
    \caption{Blimp: Comparison of five standard flight trajectories. Flight data in red, trajectory reference in blue. Rows from top to bottom are NR-based technique (labeled by the acronym NRT), NMPC, and FBL-based controller.}
    \label{fig:blimp_normal}
\end{figure*}

The FBL-based controller selects outputs $\sigma = [p_x\,p_y\,p_z\,\psi]^\top$ and inputs $u=[f_x\,f_y\,f_z\,\tau_z]^\top$. The blimp dynamics can be written in the control-affine form
\begin{equation}
    \dot{\nu}_{b/n}^b
    = N(x) \nu_{b/n}^b + F(x) + K u
\end{equation}
with matrices $N(x)$, $F(x)$ and $K$ given in~\cite[Equation 5--7]{MK-LB-SC:2024}.

Under appropriate geometric conditions, by~\cite[Theorem 1]{MK-LB-SC:2024} the blimp is input-output feedback linearizable such that the second derivative of the outputs can be expressed in canonical form as
\begin{equation}
    \ddot{\sigma}=a(x)+B(x)u \coloneqq q
\end{equation}
for state-dependent vector $a(x)\in\R^4$ and invertible matrix $B(x)\in\R^{4\times 4}$.

This system is cast as a double integrator system with a virtual input $q$ after performing a dynamic inversion of the form
\begin{equation}
    u(x) = B^{-1}(x) (q - a(x)).
\end{equation}
Now, given a reference trajectory $(r(t), \dot{r}(t), \ddot{r}(t))$ and corresponding error signal $e(t) := \sigma(t) - r(t)$, the feedback linearization controller tracks the reference via a proportional-derivative controller with feed-forward acceleration as the virtual input,
\begin{equation}
    q=-k_1 e - k_2 \dot{e} + \ddot{r}.
\end{equation}
Additionally, the feedback linearization controller is augmented with high-order CBFs that empirically mitigate the worst-case roll and pitch. For details on the implementation as a quadratic program, see~\cite{MK-LB-SC:2024}.

The nonlinear model predictive controller that we compare to is also the baseline controller in~\cite{MK-LB-SC:2024}. Given the blimp dynamics in the form $\dot{x} = f(x,u)$, let $(r(t), \dot{r}(t))$ be the reference trajectory, and let $e(t) = \big[\sigma(t)^\top - r(t)^\top\ \ \dot{\sigma}(t)^\top - \dot{r}(t)^\top\big]^\top$ be the tracking error signal and its derivative.
After discretizing the dynamics with a zero-order hold, the optimization program
\begin{align}
    u^*_k =& \underset{u_k}{\text{argmin}} \sum_{j=k}^N e_j^\top Q e_j + u_j^\top R u_j\\
    \text{s.t. }& x_{j+1} = x_j + \Delta t f(x_j, u_j)\quad j=k,\dots,N-1\nonumber\\
    & u_j \in \calU\quad j=k,\dots,N-1\nonumber
\end{align}
is repeatedly solved in CasADi\cite{CASADI}, where $\Delta t$ is the discretization step, $Q\in\R^{8\times8}$ and $R\in\R^{4\times4}$ are cost matrices, and $\calU$ represents the actuation limits. For both the blimp and quadrotor, the cost matrices were identified through empirical tuning in which candidate matrices were tested and the final selection was made to minimize the tracking error.

\subsection{Integral Control Barrier Functions for Smooth Actuator Saturation}\label{subsec:NR_ICBFs}
Integral control barrier functions (I-CBFs)~\cite{ICBF} are an extension of CBFs for dynamically defined control laws. I-CBFs can enforce forward invariance for both states and inputs by treating the input as an extended state variable. We utilize I-CBFs for smooth actuator saturation when utilizing the NR-based tracking technique. Denoting the right hand side of~\eqref{eqn:dynamic_system_NRflow_withspeedup} by $\Psi\bigl(x(t),u(t),t\bigr)$, we modify the control law to
\begin{equation}
\label{I-CBF_general_control_law}
\dot{u}(t) = \Psi\bigl(x(t),u(t),t\bigr) + {\eta}(t),
\end{equation}
where $\eta(t)$ is a minimal I-CBF intervention term that smoothly limits actuation. These developments yield a continuous-time Newton-Raphson-based dynamic control law with smooth input saturation to enhance flight safety. See~\cite{ICBF,AA-SC-ME-GN-KS-PT:2019} for more information on I-CBFs and general CBFs, respectively.

\begin{figure}
    \centering \includegraphics[width=0.63\linewidth]{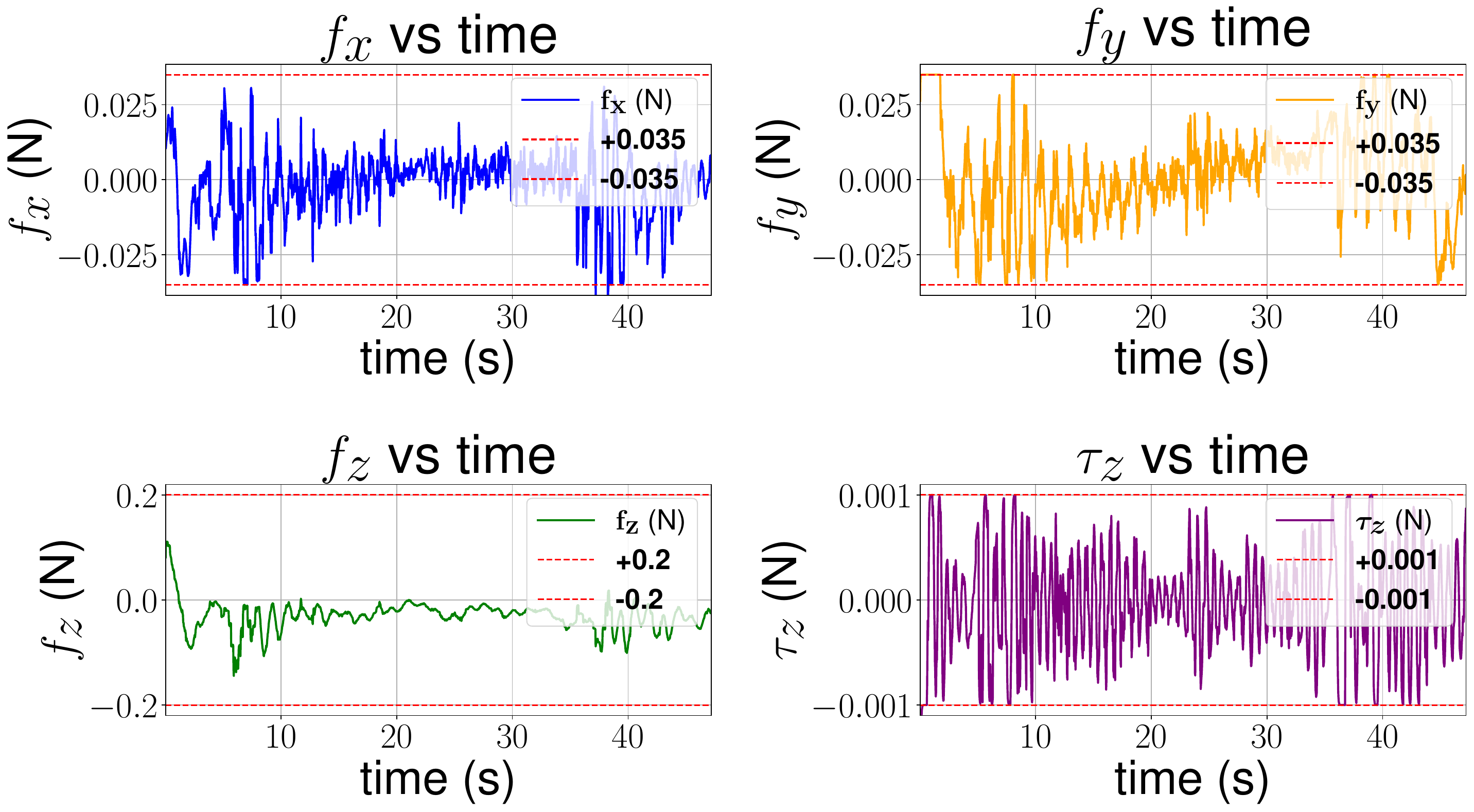}
    \caption{Force and torque inputs from the NR-based technique smoothly saturated to within safe limits (red horizontal lines) via I-CBFs.}
    \label{fig:blimp_cbfs}
\end{figure}

Figure~\ref{fig:blimp_cbfs} displays the control inputs for the blimp in terms of forces and torque about the z-axis to track the horizontal circle trajectory. These are prescribed by the NR-based technique and kept within safe limits with I-CBFs, outlined by the red horizontal lines.
This saturation is smooth in the sense that it considers the speed at which the system approaches violations of the safety constraint and begins making adjustments before the safe limits are reached, unlike hard saturation methods that abruptly clip the control input once a bound is exceeded.

Although CBFs guarantee forward invariance of the safe set in theory, practical hardware implementations can introduce violations of this property. Contributing factors include model mismatch, finite-precision computation, sensor noise, and unmodeled disturbances. Most critically, standard CBF forward invariance proofs assume continuous-time implementation, whereas digital control imposes discretization and sampling effects that may cause temporary excursions outside the safe set. However, as shown in Figure~\ref{fig:blimp_cbfs}, the inputs are always immediately pushed back into the safe region by the I-CBFs when this occurs.

\subsection{Blimp Experimental Setup and Results}
The blimp gondola is comprised of six counter-rotating motors (two vertical and four horizontal), an inertial measurement unit (IMU), a motor board, and a Raspberry Pi Zero W. IMU data as well as motion-capture data from an OptiTrack system are streamed over a WiFi connection to a Dell Precision 5570 laptop with a 12th Gen Intel Core i7-12700H $\times$ 20 CPU. The laptop first fuses IMU and motion-capture data to estimate states, and utilizes this information for control synthesis, from which appropriate motor commands are calculated. These motor commands are sent to the motor board via a base station over a non-WiFi \qty{2.4}{GHz} connection.

\begin{figure*}
    \centering
    \includegraphics[width=1.0\textwidth]{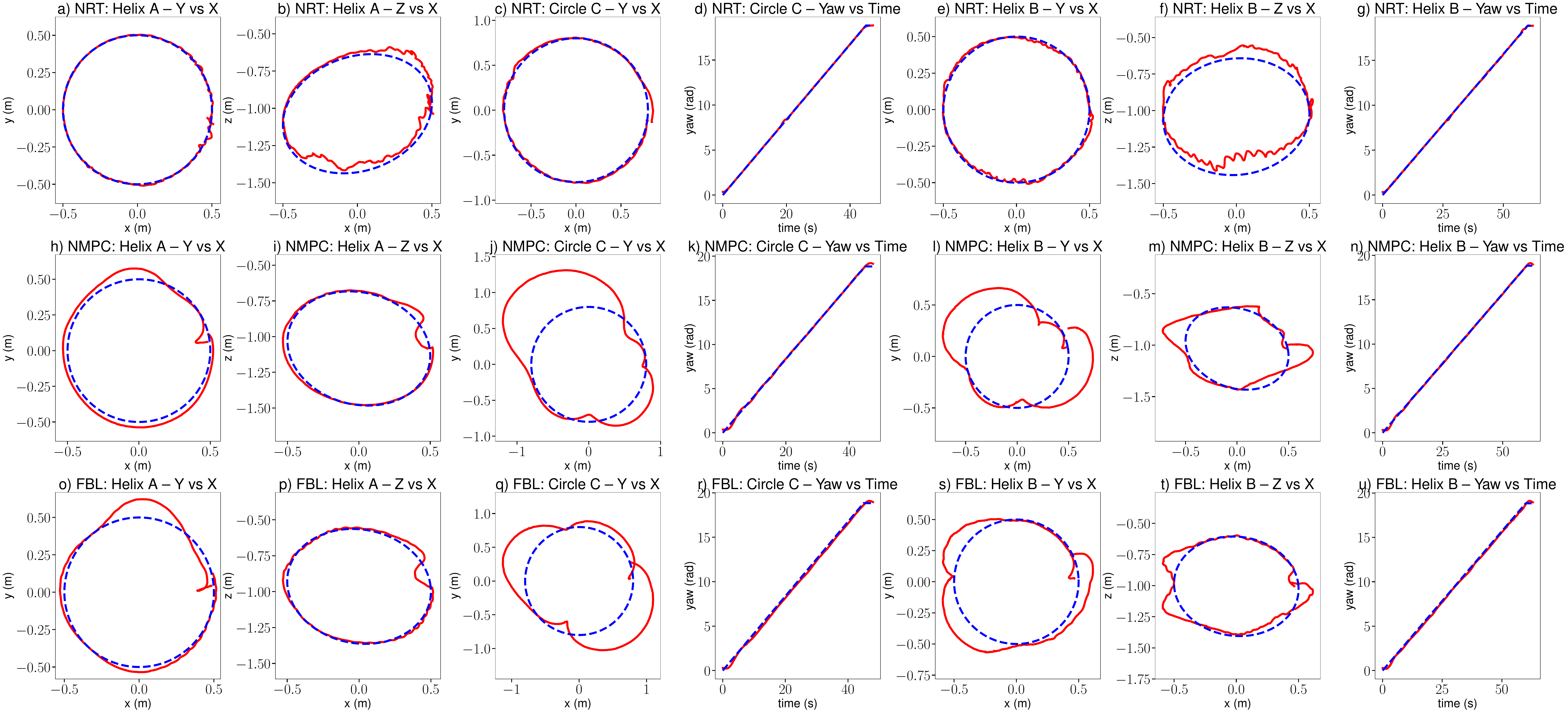}
    \caption{Blimp: Comparison of three aggressive trajectories. Flight data in red, trajectory reference in blue. Rows from top to bottom are from the NR-based technique, NMPC, and FBL-based controller.}
    \label{fig:blimp_aggressive}
\end{figure*}

We compare the NR-based technique, NMPC, and FBL-based controllers on the miniature blimp. The tracking results are plotted in Figures~\ref{fig:blimp_normal}, \ref{fig:blimp_aggressive} with error and computation time given in Tables~\ref{tab:RMSE_blimp_clipped_final}, \ref{tab:computation_time_blimp}, respectively. Across all tables, entries typeset in bold indicate the best-performing result in the corresponding performance category. For brevity, the NR-based technique is denoted by the acronym `NRT' in all figures and tables. Computation time is always measured by the real-world time taken for the function call for each controller to return its value at every iteration, measured with the \texttt{time} Python module. In all tables, bold values indicate the best performance for each trajectory.

We selected five standard trajectories for the miniature blimp. These are Horizontal Circle (Circle A), Vertical Circle (Circle B), Horizontal Lemniscate (Lemniscate A), Vertical Short Lemniscate (Lemniscate B), Vertical Tall Lemniscate (Lemniscate C).
We additionally selected three aggressive trajectories for the blimp: Regular Helix (Helix A), Yawing Helix (Helix B), and Yawing Horizontal Circle (Circle C). For brevity, we sometimes shorten trajectory names or use the abbreviated names in parentheses (e.g., Horizontal Circle as Circle A).

\begin{table}[h]
  \centering
  \scriptsize
  \setlength{\tabcolsep}{4pt}
  \renewcommand{\arraystretch}{0.95}
  \caption{Blimp - NR-based technique vs. NMPC vs. FBL RMSE}
  \label{tab:RMSE_blimp_clipped_final}
  \resizebox{.50\linewidth}{!}{
  \begin{tabular}{|c|c|c|c|}
    \hline
    \textbf{Trajectory} & \textbf{NRT [m]} & \textbf{NMPC [m]} & \textbf{FBL [m]}\\
    \hline
    Circle A      & \textbf{0.07866} & 0.14075 & 0.19332 \\
    \hline
    Circle B      & \textbf{0.05137} & 0.10991 & 0.14255 \\
    \hline
    Lemniscate A  & \textbf{0.10441} & 0.18517 & 0.22955 \\
    \hline
    Lemniscate B  & \textbf{0.05389} & 0.10291 & 0.14565 \\
    \hline
    Lemniscate C  & \textbf{0.05594} & 0.11793 & 0.13073 \\
    \hline
    Helix A       & \textbf{0.07187} & 0.07732 & 0.11291 \\
    \hline
    Helix B       & \textbf{0.08767} & 0.21534 & 0.22001 \\
    \hline
    Circle C      & \textbf{0.10124} & 0.40327 & 0.38360 \\
    \hline
  \end{tabular}
  }
\end{table}

\vspace{.1in}

\begin{table}[h]
  \centering
  \scriptsize
  \setlength{\tabcolsep}{4pt}
  \renewcommand{\arraystretch}{0.95}
  \caption{Blimp - Computation Time Comparison}
  \label{tab:computation_time_blimp}
  \resizebox{.57\linewidth}{!}{
  \begin{tabular}{|c|c|c|c|}
    \hline
    \textbf{Trajectory} & \textbf{NRT [ms]} & \textbf{NMPC [ms]} & \textbf{FBL [ms]}\\
    \hline
    Circle A      & \textbf{0.84 $\pm$ 0.046} & 58.67 $\pm$ 15.67 & 10.42 $\pm$ 9.69 \\
    \hline
    Circle B      & \textbf{0.82 $\pm$ 0.031} & 51.47 $\pm$ 12.23 & 10.86 $\pm$ 10.43 \\
    \hline
    Lemniscate A  & \textbf{0.82 $\pm$ 0.019} & 52.67 $\pm$ 12.49 & 10.51 $\pm$ 10.66 \\
    \hline
    Lemniscate B  & \textbf{0.87 $\pm$ 0.23}  & 54.18 $\pm$ 6.84  & 10.84 $\pm$ 10.12 \\
    \hline
    Lemniscate C  & \textbf{0.86 $\pm$ 0.018} & 54.84 $\pm$ 11.93 & 10.72 $\pm$ 10.42 \\
    \hline
    Helix A       & \textbf{0.94 $\pm$ 0.041} & 53.15 $\pm$ 6.78  & 9.89 $\pm$ 9.49 \\
    \hline
    Helix B       & \textbf{0.95 $\pm$ 0.022} & 60.76 $\pm$ 12.78 & 10.27 $\pm$ 10.05 \\
    \hline
    Circle C      & \textbf{0.96 $\pm$ 0.034} & 73.49 $\pm$ 20.92 & 10.18 $\pm$ 9.88 \\
    \hline
  \end{tabular}
  }
\end{table}

\subsection{Blimp Analysis and Discussion}

The NR-based tracking technique is an order of magnitude faster than the FBL-based controller, which is itself about half-an order of magnitude faster than the NMPC controller, see Table \ref{tab:computation_time_blimp}. We note that the FBL-based controller calls on CBFs to mitigate oscillations in its internal dynamics, which factor into its computation time.

Given a prescribed control update rate of \qty{40}{Hz}, the NR-based and FBL-based controllers consistently meet the corresponding \qty{25}{ms} deadline necessary for low-level control of our particular blimp platform. However, due to computational constraints, the NMPC controller does not meet its deadline and consequently tracking performance is degraded. Note that when performing a simulation with real-time computational requirements removed and a \qty{3}{s} lookahead, the NMPC controller achieves the lowest tracking error; however, this may not be realizable within realistic computational constraints.

We highlight an additional advantage of the
NR-based technique. For both the FBL-based controller as well as NMPC to achieve competitive tracking performance, we found it was necessary to enhance them with reference trajectory derivative information. On the other hand, the NR-based technique was only given the desired trajectory with no derivative information. Nevertheless, the RMSE for the NR-based technique was between one-half and one-quarter of that of the other controllers, as seen in Table~\ref{tab:RMSE_blimp_clipped_final}.

\section{Quadrotor}\label{sec:experiments_quadrotor}
Quadrotor research is led by work on small, agile indoor racing drones with significant onboard computing capabilities. In particular, quadrotors small enough to be classified as Micro-unmanned Aerial Vehicles (MAVs) of a size between \qtyrange{0.1}{0.5}{m} in diagonal length and \qtyrange{0.1}{0.5}{kg} of mass \cite{mUAV_challenges_and_opportunities, kushleyev_towards_swarm_of_agile_microquads} are commonly used.

While smaller quadrotors are desirable for upkeep and for multi-agent research in small flight spaces, the primary reason for their expansive use lies in the physics of quadrotor flight. A quadrotor's translational motion is controlled by tilting its body, meaning rotational dynamics directly influence translational agility. Consequently, because angular acceleration is inversely proportional to the size of the quadrotor \cite{kushleyev_towards_swarm_of_agile_microquads}, MAVs feature prominently in present-day research.

\begin{figure}[h]
    \centering
    \includegraphics[width=0.2\textwidth]{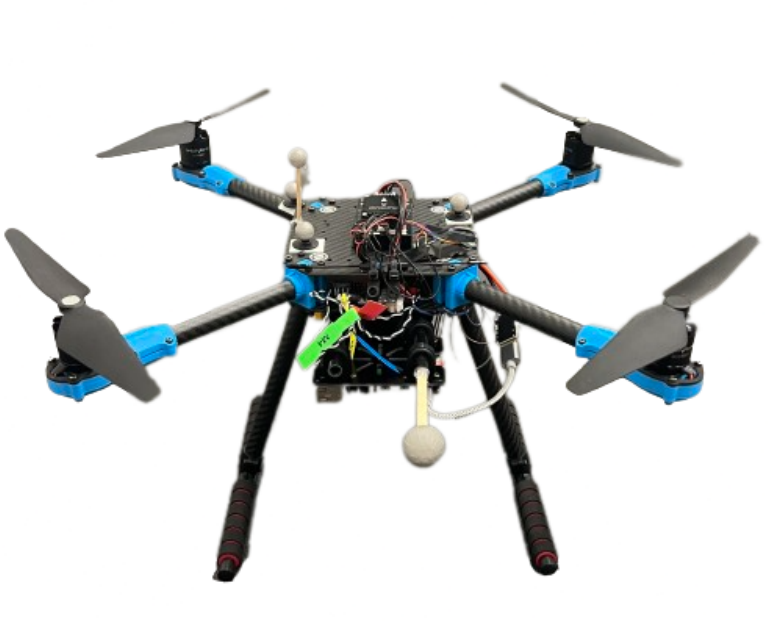}
    \caption{The Holybro X500 V2 quadrotor used for hardware experiments. It is fitted with tracking markers, a radio receiver, and an on-board Raspberry Pi with ROS2 for control computations. The quadrotor weighs \qty{2.1}{kg}, measures \qty{0.6}{m} diagonally, and \qty{0.25}{m} vertically.}
    \label{fig:my_quadrotor_image}
\end{figure}

However, while less maneuverable, medium-sized and larger quadrotors have the inertia necessary to withstand unfavorable real-world weather conditions and can carry larger payloads~\cite{uav_handbook, kushleyev_towards_swarm_of_agile_microquads}. This makes research exploiting their strengths and addressing their limitations both useful for real-world deployment and less commonly addressed in work that focuses on small racing drones. In contrast to the dominant focus on MAV agility, our work investigates the complementary challenges posed by larger platforms. We begin with the challenges of computational constraints and limited vehicle agility, and seek to maximize trajectory speed and tracking accuracy despite these constraints. The quadrotor used in our flight experiments is depicted in Figure~\ref{fig:my_quadrotor_image}.

\subsection{Dynamic Model and Tracking Controllers for the Quadrotor}\label{subsec:quad_model}

\begin{figure*}
    \centering
    \includegraphics[width=1.0\linewidth]{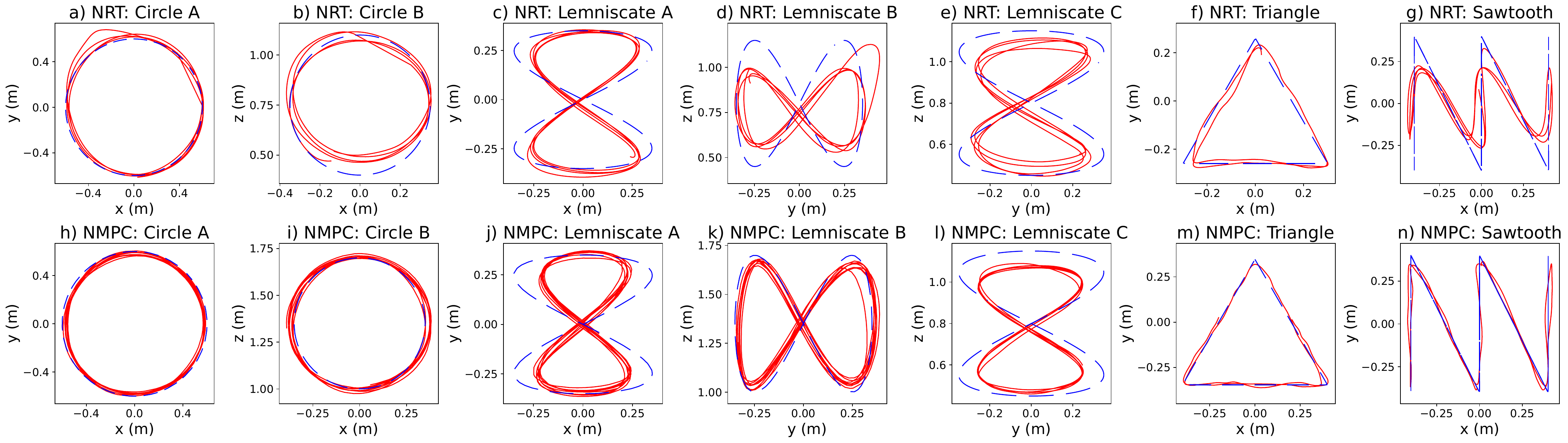}
    \caption{Quadrotor: Flight data in red, trajectory reference in blue. Rows from top to bottom are the NR-based technique and NMPC.}
    \label{fig:quad_standard_2x}
\end{figure*}

We model our quadrotor as a 9-state, 4-input system. We take the system state as
\begin{align}
{x} &= \begin{bmatrix}
        p_x & p_y & p_z & V_x & V_y & V_z & \phi & \theta & \psi
    \end{bmatrix}^{\top}\in\mathbb{R}^9
\end{align}
where $(p_x, p_y, p_z)$ is the system's 3D position in a fixed north, east, down (NED) world frame, $(V_x, V_y, V_z)$ is the system's velocity within the frame, and $(\phi, \theta, \psi)$ are Euler angles describing the orientation of the system within the same coordinate framework.

The system input is
\(
u = \begin{bmatrix} u_\tau, \; u_p, \; u_q, \; u_r \end{bmatrix}^{\top} \in \mathbb{R}^4,
\)
where $u_\tau$ is the net thrust through the quadrotor's body-frame $z$-axis and
$\omega^{b} = (u_p, u_q, u_r)$ are the angular body rates of the quadrotor. The nonlinear dynamics are
\begin{equation}
\label{eqn:quad_nonlindyn}
    \begin{aligned}
        &(\dot{p}_x,\dot{p}_y,\dot{p}_z) = (V_x, V_y, V_z),\\
        &\dot{V}_x = -\tfrac{u_\tau}{m} \Bigl(\sin\phi \sin\psi + \cos\phi \cos\psi \sin\theta \Bigr),\\
        &\dot{V}_y = -\tfrac{u_\tau}{m} \Bigl(\cos\phi \sin\psi \sin\theta - \cos\psi \sin\phi \Bigr),\\
        &\dot{V}_z = g - \tfrac{u_\tau}{m} \Bigl(\cos\phi \cos\theta \Bigr),\\
        &(\dot{\phi},\dot{\theta},\dot{\psi}) = T \, \omega^{b},
    \end{aligned}
\end{equation}
where $T$ is the matrix for angular velocity transformations \cite{Sabatino860649}, namely
\[
T =
\begin{bmatrix}
1 & \sin\phi \tan\theta & \cos\phi \tan\theta \\
0 & \cos\phi & -\sin\phi \\
0 & \tfrac{\sin\phi}{\cos\theta} & \tfrac{\cos\phi}{\cos\theta}
\end{bmatrix}.
\]

For the quadrotor, we compare the NR-based technique against a compiled NMPC algorithm built with the \texttt{acados}~\cite{acados} fast embedded optimal control toolbox.
Across all methods for both the quadrotor and the blimp, we employ conservative tuning approaches that favor out-of-the-box configurations and minimal solver-level customization. For example, although more specialized adjustments to NMPC---such as modified termination criteria, customized tolerances, or tailored warm-start schemes---could reduce computation times for certain trajectories, such enhancements typically require substantial modification and may obscure the source of the performance gains. In contrast, the NR-based technique exhibits predictable behavior under minimal tuning. For these reasons, we evaluate all methods under well-tuned yet conservative numerical configurations.

In particular, the NMPC scheme deployed for the quadrotor is configured using a Real-Time Iteration (RTI) approach~\cite{MD_HB_JS_2005}, which performs a single SQP linearization and QP solve at each sampling instant, rather than iterating to convergence.
While this reduces the computational burden, the average solve time still exceeds one sampling period on the onboard computer (see Table~\ref{tab:quad_computation_time_comparison}). This is a well-known concern in modern NMPC implementations. To handle this, we follow a computational delay compensation strategy~\cite{RF_2005_Thesis}
to ensure consistent command timing despite variable solver execution times. In addition, implementing the tracking error within the CasADi quadrotor model, rather than within the \texttt{acados} NMPC formulation, was found to improve yaw tracking performance. Collectively, these implementation choices were necessary to achieve stable tracking across all trajectories, avoiding undesirable behaviors such as shaking and jitter.

\subsection{Quadrotor Experimental Setup and Results}\label{subsec:quad_setup}

Leveraging the cascaded control architecture of the PX4 flight stack~\cite{PX4_Controller_Diagrams}, we send rate control commands to the innermost loop of the stack. This control structure makes the system highly sensitive to small input variations, requiring continuous micro-adjustments. Consequently, tracking controllers deployed on such a system must publish commands at rates of at least \qty{100}{\hertz} to ensure flight stability.

In our hardware experiments, we use a {Holybro X500} quadrotor equipped with a {Pixhawk 6X} flight controller as the flight management unit. Onboard control computations are performed using a Raspberry Pi 4 Model B transmitting control commands to the flight controller via a serial connection.
Communication between the Raspberry Pi and the flight controller is facilitated through the ROS 2 API available within the PX4 flight stack. The $\mu\text{XRCE-DDS}$ middleware serves as a bridge between PX4's native uORB topics and ROS 2 topics. For motion capture, we employ the OptiTrack system and use a relay to convert OptiTrack messages into visual odometry messages. These are subsequently fused with the flight controller's onboard sensor data via the flight stack's extended Kalman filter.

Despite the significant differences between the blimp and quadrotor platforms, the NR-based tracking technique exhibits a modular structure that supports cross-platform deployment. Porting the controller requires (i) substituting the appropriate vehicle dynamics within the internal prediction model, and (ii) updating the input evolution law $u_{{\rm next}} = u + \dot{u}\,\Delta t$ to match the platform's control-rate and actuation conventions. Additionally, incorporating platform-specific safe control limits into the I-CBF constraints ensures compatibility with each vehicle's actuator capabilities and desired flight envelope. The remaining effort consists of tuning the speedup factor $\alpha$ and lookahead horizon $T$. While these steps are not trivial, the controller's structure makes the adaptation process systematic and relatively straightforward.

Due to the higher maneuverability of quadrotors compared to blimps, more complex trajectories are possible. We selected a total of eight trajectories for the quadrotor, with variations accounting for the total of twenty-two trajectories. We first test the five standard trajectories as explained in Section \ref{sec:experiments_blimp}. We also introduce a new set of trajectories, called ``aggressive'' trajectories. These include performing the five standard trajectories at double their original speed, as well as three new trajectories: Helix, Sawtooth, and Triangle. We also introduce variants that incorporate both path tracking and simultaneous yaw reference tracking for the standard trajectories. Lastly, some of the aggressive trajectories are also tested at double their original travel speeds as well as double the original yawing speeds.

We again utilize the five standard trajectories from the Blimp experiments: Horizontal Circle (Circle A), Vertical Circle (Circle B), Horizontal Lemniscate (Lemniscate A), Vertical Short Lemniscate (Lemniscate B), and Vertical Tall Lemniscate (Lemniscate C). We also include the aggressive trajectories: Regular Helix (Helix A), Yawing Helix (Helix B), and Yawing Horizontal Circle (Circle C). Due to the higher maneuverability of quadrotors compared to blimps, we implement two additional aggressive trajectories:
Sawtooth and Triangle. Figure~\ref{fig:quad_standard_2x} plots tracking results and provides a visual for the trajectories which can be plotted in two dimensions.\footnote{As for the case of the blimp, we use the acronym ``NRT'' for ``NR Technique'' in Figure 8 and Tables III and IV.}

Comparative results of the five standard trajectories and the five aggressive trajectories are summarized in
Tables~\ref{tab:quad_RMSE_merged} and~\ref{tab:quad_computation_time_comparison}, providing quantitative differences between the NR-based technique and the NMPC approach.

\begin{table}
  \centering
  \scriptsize
  \setlength{\tabcolsep}{4pt}
  \renewcommand{\arraystretch}{0.95}
  \caption{Quadrotor - NR-based technique vs. NMPC, RMSE Comparison (Clipped vs. Non-Clipped Transients)}
  \resizebox{.55\linewidth}{!}{
  \label{tab:quad_RMSE_merged}
  \begin{tabular}{|c|cc|cc|}
    \hline
    \textbf{Trajectory} & \multicolumn{2}{c|}{\textbf{Clipped Transients}} & \multicolumn{2}{c|}{\textbf{With Transients}} \\
    \cline{2-5}
    & \textbf{NRT [m]} & \textbf{NMPC [m]} & \textbf{NRT [m]} & \textbf{NMPC [m]} \\
    \hline
    Circle A      & 0.10695 & \textbf{0.03932} & 0.12266 & \textbf{0.10653} \\
    \hline
    Circle B      & 0.11887 & \textbf{0.03246} & \textbf{0.12276} & 0.15162 \\
    \hline
    Lemniscate A  & 0.12328 & \textbf{0.09738} & 0.12249 & \textbf{0.11311} \\
    \hline
    Lemniscate B  & 0.15915 & \textbf{0.04683} & 0.15203 & \textbf{0.10479} \\
    \hline
    Lemniscate C  & 0.14554 & \textbf{0.09938} & \textbf{0.14532} & 0.17329 \\
    \hline
    Helix A       & 0.14879 & \textbf{0.13327} & \textbf{0.14867} & 0.16865 \\
    \hline
    Helix B       & 0.19540 & \textbf{0.14977} & 0.18851 & \textbf{0.18044} \\
    \hline
    Circle C      & 0.17398 & \textbf{0.14782} & \textbf{0.17073} & 0.17715 \\
    \hline
    Sawtooth      & 0.04014 & \textbf{0.02331} & \textbf{0.04505} & 0.06032 \\
    \hline
    Triangle      & 0.05979 & \textbf{0.01803} & 0.08136 & \textbf{0.08387} \\
    \hline
  \end{tabular}}
\end{table}

\subsection{Quadrotor Analysis and Discussion}\label{subsec:quad_analysis}
We note that the
flight trajectories are designed
such that the quadrotor
hovers at the origin for five seconds before initiating each maneuver, and returns
to the origin to hover again at the end. These transitions introduce discontinuities in the
reference,
resulting in a transient portion of flight.
To account for
this, we report performance both with and without the transient segment, using data from the best runs of each method on each trajectory to enable fair comparisons that reflect the most favorable performance of each approach.

If we clip these transient portions of flight to consider strictly the trajectory tracking abilities of the NMPC and NR-based techniques, as seen in the ``Clipped Transients'' column of Table~\ref{tab:quad_RMSE_merged}, NMPC outperforms the NR-based technique in terms of RMSE for all successful flights. We note again that without computational delay compensation strategies, NMPC was susceptible to shaking behavior and crashed while attempting Helix B and Circle C. Further, as noted previously, the tracking performance of both NMPC and the NR-based technique can be degraded when the initial position and initial target are far apart. When the transient flight portions are considered (``With Transients'' column in~Table~\ref{tab:quad_RMSE_merged}), the NR-based technique is capable of outperforming the NMPC technique on some trajectories.

\begin{table}
  \centering
  \scriptsize
  \setlength{\tabcolsep}{4pt}
  \renewcommand{\arraystretch}{0.95}
  \caption{Quadrotor - Average Computation Time Comparison for the NR-based technique and NMPC}
  \label{tab:quad_computation_time_comparison}
  \resizebox{.42\linewidth}{!}{
  \begin{tabular}{|c|c|c|}
    \hline
    \textbf{Trajectory} & \textbf{NRT [ms]} & \textbf{NMPC [ms]} \\
    \hline
    Circle A       & \textbf{2.381 $\pm$ 0.206} & 13.036 $\pm$ 0.173 \\
    \hline
    Circle B       & \textbf{2.426 $\pm$ 0.201} & 13.352 $\pm$ 0.139 \\
    \hline
    Lemniscate A   & \textbf{2.438 $\pm$ 0.239} & 13.088 $\pm$ 0.092 \\
    \hline
    Lemniscate B   & \textbf{2.486 $\pm$ 0.200} & 13.288 $\pm$ 0.246 \\
    \hline
    Lemniscate C   & \textbf{2.397 $\pm$ 0.200} & 9.216 $\pm$ 5.646 \\
    \hline
    Helix A        & \textbf{2.480 $\pm$ 0.222} & 13.307 $\pm$ 0.196 \\
    \hline
    Helix B        & \textbf{2.471 $\pm$ 0.271} & 13.294 $\pm$ 0.110 \\
    \hline
    Circle C       & \textbf{2.454 $\pm$ 0.255} & 13.705 $\pm$ 0.222 \\
    \hline
    Sawtooth       & \textbf{2.507 $\pm$ 0.254} & 13.329 $\pm$ 0.140 \\
    \hline
    Triangle       & \textbf{2.417 $\pm$ 0.196} & 13.182 $\pm$ 0.291 \\
    \hline
  \end{tabular}}
\end{table}

Next, Table~\ref{tab:quad_computation_time_comparison} shows that the computation times for NMPC are generally between 5.32 and 5.59 times larger than those of the NR-based technique. Moreover, while the computation times of the NR-based technique are largely independent of the trajectory being tracked, the computation time for NMPC with full SQP solves exhibited significant variability. For the SQP-RTI configuration reported in Table~\ref{tab:quad_computation_time_comparison}, this variability is less pronounced, with the exception of Lemniscate~C.

Noting that Helix B and Circle C differ from Helix A and Circle A by including yaw variations, it was found that for full SQP solves, this seems to increase the required computing times for NMPC beyond the \qty{100}{Hz} rate constraint. Consequently, each attempt to track the yawing trajectories led to crashes. Even on the standard trajectories with SQP-RTI, NMPC computation on average takes more than the \qty{10}{ms} allotted for each iteration. The decoupled compute-and-publish architecture described above circumvents this issue by ensuring that command publication remains consistently timed at \qty{100}{\hertz} regardless of solver runtime. This allows NMPC to perform all trajectories, often achieving lower RMSE than the NR-based controller. Lastly, average energy expenditures across all tracking controllers for both quadrotor and blimp are shown in Table~\ref{tab:energy_expenditure_comparison}. The NR-based technique has the lowest CPU energy expenditure across all controllers used for both the blimp and the quadrotor.

\begin{table}
  \centering
  \scriptsize
  \setlength{\tabcolsep}{4pt}
  \caption{Average CPU Energy Expenditure Comparison per Method on Blimp and Quadrotor}
  \label{tab:energy_expenditure_comparison}
  \resizebox{.43\linewidth}{!}{

  \begin{tabular}{|c|c|}
    \hline
    \textbf{Method} & \textbf{Avg. CPU Energy Expenditure [\boldmath$\mu$J]} \\
    \hline
    Blimp NRT   & \boldmath{$1.25 \times 10^4 \pm 4.29 \times 10^3$} \\
    \hline
    Blimp MPC  & $2.04 \times 10^5 \pm 3.80 \times 10^4$ \\
    \hline
    Blimp FBL  & $3.06 \times 10^4 \pm 2.22 \times 10^4$ \\
    \hline
    Quad NRT & \boldmath{$1.73 \times 10^{4} \pm 7.78 \times 10^{3}$} \\
    \hline
    Quad MPC & $6.15 \times 10^{4} \pm 2.41 \times 10^{4}$ \\
    \hline
  \end{tabular}}
\end{table}

\section{Conclusions}\label{sec:conclusion}
This work compares the proposed Newton-Raphson-based tracking control technique against well-established alternatives from the literature on two aerial hardware platforms under realistic deployment and computational constraints. The comparison is based on three primary metrics: tracking accuracy, computation time, and CPU energy expenditure. The proposed NR-based controller
outperforms the FBL-based control method from the blimp literature across all three metrics. Compared to the more computationally demanding
NMPC controller, the NR-based technique demonstrates shorter computing times and less CPU energy usage while maintaining competitive tracking accuracy across most trajectories, and in some cases, achieving greater accuracy.

When given ample time, computational power, and agile hardware like micro-UAVs, NMPC achieves exceptional performance for aggressive trajectory tracking, as is well documented in the literature. However, its suitability may be limited in deployment scenarios where computational constraints prevent meeting real-time requirements for low-level control. As demonstrated in this study, the NR-based technique provides a lightweight, adaptable, and theoretically grounded alternative, making it a viable choice for resource-constrained applications across a range of systems.

Future work on the NR-based tracking technique may explore enhancements to the prediction mechanism, including the use of adaptive or learning-based predictors, as well as uncertainty-aware prediction models. In addition, the speedup parameter \(\alpha\) has thus far been restricted to a tuned constant vector in all existing implementations of the NR-based technique. Future research could investigate adaptive schemes for \(\alpha\), as well as generalized formulations in which \(\alpha\) is promoted to a full matrix of speedup parameters, thereby incorporating cross-terms and enabling an examination of their impact on tracking performance and stability.

\bibliographystyle{plain}
\bibliography{2_arXiv_Version/references}

@phdthesis{RF_2005_Thesis,
  author       = {Rolf Findeisen},
  title        = {Nonlinear Model Predictive Pontrol : A Sampled Data Feedback Perspective},
  school       = {University of Stuttgart},
  year         = {2005},
  type         = {PhD thesis},
  url          = {https://elib.uni-stuttgart.de/items/e865a9fa-0cbf-47e8-8d8c-916debc64ec4}
}

@article{MR_2023,
   title={Model predictive control of non-holonomic systems: Beyond differential-drive vehicles},
   volume={152},
   ISSN={0005-1098},
   url={http://dx.doi.org/10.1016/j.automatica.2023.110972},
   DOI={10.1016/j.automatica.2023.110972},
   journal={Automatica},
   publisher={Elsevier BV},
   author={Rosenfelder, Mario and Ebel, Henrik and Krauspenhaar, Jasmin and Eberhard, Peter},
   year={2023},
   month=jun, pages={110972} }

@article{MD_HB_JS_2005,
author = {Diehl, Moritz and Bock, Hans Georg and Schl\"{o}der, Johannes P.},
title = {A Real-Time Iteration Scheme for Nonlinear Optimization in Optimal Feedback Control},
journal = {SIAM Journal on Control and Optimization},
volume = {43},
number = {5},
pages = {1714-1736},
year = {2005},
doi = {10.1137/S0363012902400713},
URL = {eprint = https://doi.org/10.1137/S0363012902400713}
}

@article{wardi_ijrnc_2019,
author = {Wardi, Y. and Seatzu, C. and Cortés, J. and Egerstedt, M. and Shivam, S. and Buckley, I.},
title = {Tracking control by the Newton–Raphson method with output prediction and controller speedup},
journal = {International Journal of Robust and Nonlinear Control},
volume = {34},
number = {1},
pages = {397-422},
doi = {https://doi.org/10.1002/rnc.6976},
url = {https://onlinelibrary.wiley.com/doi/abs/10.1002/rnc.6976},
eprint = {https://onlinelibrary.wiley.com/doi/pdf/10.1002/rnc.6976},
year = {2024}
}

@INPROCEEDINGS{8264633,
  author={Wardi, Y. and Seatzu, C. and Egerstedt, M. and Buckley, I.},
  booktitle={2017 IEEE 56th Annual Conference on Decision and Control (CDC)}, 
  title={Performance regulation and tracking via lookahead simulation: Preliminary results and validation}, 
  year={2017},
  volume={},
  number={},
  pages={6462-6468},
  doi={10.1109/CDC.2017.8264633}}

@INPROCEEDINGS{mypaper,
  author={Morales-Cuadrado, Evanns and Llanes, Christian and Wardi, Yorai and Coogan, Samuel},
  booktitle={2024 American Control Conference (ACC)}, 
  title={Newton-Raphson Flow for Aggressive Quadrotor Tracking Control}, 
  year={2024},
  volume={},
  number={},
  pages={3879-3884},
  doi={10.23919/ACC60939.2024.10644692}}

@misc{acados,
      title={acados: a modular open-source framework for fast embedded optimal control}, 
      author={Robin Verschueren and Gianluca Frison and Dimitris Kouzoupis and Jonathan Frey and Niels van Duijkeren and Andrea Zanelli and Branimir Novoselnik and Thivaharan Albin and Rien Quirynen and Moritz Diehl},
      year={2020},
      eprint={1910.13753},
      archivePrefix={arXiv},
      primaryClass={math.OC},
      url={https://arxiv.org/abs/1910.13753}, 
}

@ARTICLE{ICBF,
  author={Ames, Aaron D. and Notomista, Gennaro and Wardi, Yorai and Egerstedt, Magnus},
  journal={IEEE Control Systems Letters}, 
  title={Integral Control Barrier Functions for Dynamically Defined Control Laws}, 
  year={2021},
  volume={5},
  number={3},
  pages={887-892},
  doi={10.1109/LCSYS.2020.3006764}}

@ARTICLE{nmpc_faulttolerant_scaramuzza,
  author={Nan, Fang and Sun, Sihao and Foehn, Philipp and Scaramuzza, Davide},
  journal={IEEE Robotics and Automation Letters}, 
  title={Nonlinear MPC for Quadrotor Fault-Tolerant Control}, 
  year={2022},
  volume={7},
  number={2},
  pages={5047-5054},
  doi={10.1109/LRA.2022.3154033}}

@Inbook{uav_handbook,
author="Powers, Caitlin
and Mellinger, Daniel
and Kumar, Vijay",
title="Quadrotor Kinematics and Dynamics",
bookTitle="Handbook of Unmanned Aerial Vehicles",
year="2015",
publisher="Springer Netherlands",
address="Dordrecht",
pages="307--328",
isbn="978-90-481-9707-1",
doi="10.1007/978-90-481-9707-1_71",
url="https://doi.org/10.1007/978-90-481-9707-1_71"
}

@article{realtimeMPC_2014,
title = {Real-time Model Predictive Control for Quadrotors},
journal = {IFAC Proceedings Volumes},
volume = {47},
number = {3},
pages = {11773-11780},
year = {2014},
note = {19th IFAC World Congress},
issn = {1474-6670},
doi = {https://doi.org/10.3182/20140824-6-ZA-1003.00203},
url = {https://www.sciencedirect.com/science/article/pii/S1474667016434890},
author = {Moses Bangura and Robert Mahony}
}

@INPROCEEDINGS{nmpc_so3,
  author={Kamel, Mina and Alexis, Kostas and Achtelik, Markus and Siegwart, Roland},
  booktitle={2015 IEEE Conference on Control Applications (CCA)}, 
  title={Fast nonlinear model predictive control for multicopter attitude tracking on SO(3)}, 
  year={2015},
  volume={},
  number={},
  pages={1160-1166},
  doi={10.1109/CCA.2015.7320769}}

@ARTICLE{scaramuzza_nmpc_dfbc,
  author={Sun, Sihao and Romero, Angel and Foehn, Philipp and Kaufmann, Elia and Scaramuzza, Davide},
  journal={IEEE Transactions on Robotics}, 
  title={A Comparative Study of Nonlinear MPC and Differential-Flatness-Based Control for Quadrotor Agile Flight}, 
  year={2022},
  volume={38},
  number={6},
  pages={3357-3373},
  doi={10.1109/TRO.2022.3177279}}

@article{kushleyev_towards_swarm_of_agile_microquads,
  title={Towards a swarm of agile micro quadrotors},
  author={Kushleyev, Alex and Mellinger, Daniel and Powers, Caitlin and Kumar, Vijay},
  journal={Autonomous Robots},
  volume={35},
  number={4},
  pages={287--300},
  year={2013},
  publisher={Springer}
}

@ARTICLE{mUAV_challenges_and_opportunities,
  author={Liu, Xu and Chen, Steven W. and Nardari, Guilherme V. and Qu, Chao and Cladera, Fernando and Taylor, Camillo J. and Kumar, Vijay},
  journal={IEEE Micro}, 
  title={Challenges and Opportunities for Autonomous Micro-UAVs in Precision Agriculture}, 
  year={2022},
  volume={42},
  number={1},
  pages={61-68},
  doi={10.1109/MM.2021.3134744}}

@mastersthesis{Sabatino860649,
   author = {Sabatino, Francesco},
   institution = {KTH, Automatic Control},
   pages = {61},
   school = {KTH, Automatic Control},
   title = {Quadrotor control: modeling, nonlinearcontrol design, and simulation},
   series = {EES Examensarbete / Master Thesis},
   year = {2015}
}

@article{scaramuzza_l1_nmpc,
   title={Performance, Precision, and Payloads: Adaptive Nonlinear MPC for Quadrotors},
   volume={7},
   ISSN={2377-3774},
   url={http://dx.doi.org/10.1109/LRA.2021.3131690},
   DOI={10.1109/lra.2021.3131690},
   number={2},
   journal={IEEE Robotics and Automation Letters},
   publisher={Institute of Electrical and Electronics Engineers (IEEE)},
   author={Hanover, Drew and Foehn, Philipp and Sun, Sihao and Kaufmann, Elia and Scaramuzza, Davide},
   year={2022},
   month=apr, pages={690–697} }

@INPROCEEDINGS{NN_Pred_1,
  author={Niu, K. and Wardi, Y. and Abdallah, C. T. and Hayajneh, M.},
  booktitle={2022 American Control Conference (ACC)}, 
  title={A Model-free Tracking Controller Based on the Newton-Raphson Method and Feedforward Neural Networks}, 
  year={2022},
  volume={},
  number={},
  pages={3254-3259},
  doi={10.23919/ACC53348.2022.9867840}}

@misc{PX4_Controller_Diagrams,
  author        = "PX4",
  title         = "Controller Diagrams",
  year          = "2023",
  howpublished  = "\url{https://docs.px4.io/main/en/flight_stack/controller_diagrams.html}",
  note          = "[Online; accessed 2023-09-21]",
  organization  = "Dronecode Project"
}

@INPROCEEDINGS{AA-SC-ME-GN-KS-PT:2019,
  author={Ames, Aaron D. and Coogan, Samuel and Egerstedt, Magnus and Notomista, Gennaro and Sreenath, Koushil and Tabuada, Paulo},
  booktitle={2019 18th European Control Conference (ECC)}, 
  title={Control Barrier Functions: Theory and Applications}, 
  year={2019},
  volume={},
  number={},
  pages={3420-3431},
  doi={10.23919/ECC.2019.8796030}}

@article{CASADI,
author = {Andersson, Joel and Gillis, Joris and Horn, Greg and Rawlings, James and Diehl, Moritz},
year = {2018},
month = {07},
pages = {},
title = {CasADi: a software framework for nonlinear optimization and optimal control},
volume = {11},
journal = {Mathematical Programming Computation},
doi = {10.1007/s12532-018-0139-4}
}

@INPROCEEDINGS{shivam_ecc2019_autonomous_vehicles,
  author    = {Shivam, Shashwat and Buckley, Ian and Wardi, Yorai and Seatzu, Carla and Egerstedt, Magnus},
  booktitle = {2019 European Control Conference (ECC)},
  title     = {Tracking Control by the Newton-Raphson Flow: Applications to Autonomous Vehicles},
  year      = {2019},
  pages     = {1562--1567},
  doi       = {10.23919/ECC.2019.8796157}
}

@ARTICLE{shivam_ifac2020_intersection_traffic,
  author  = {Shivam, Shashwat and Wardi, Yorai and Egerstedt, Magnus and Kanellopoulos, Aris and Vamvoudakis, Kyriakos G.},
  title   = {Intersection-Traffic Control of Autonomous Vehicles using Newton-Raphson Flows and Barrier Functions},
  journal = {IFAC-PapersOnLine},
  year    = {2020},
  volume  = {53},
  number  = {2},
  pages   = {15733--15738},
  doi     = {10.1016/j.ifacol.2020.12.054}
}

@ARTICLE{falcone_tcst_2007_active_steering,
  author  = {Falcone, Paolo and Borrelli, Francesco and Asgari, Jahan and Tseng, H. Eric and Hrovat, Davor},
  title   = {Predictive Active Steering Control for Autonomous Vehicle Systems},
  journal = {IEEE Transactions on Control Systems Technology},
  year    = {2007},
  volume  = {15},
  number  = {3},
  pages   = {566--580},
  doi     = {10.1109/TCST.2007.894653}
}

@ARTICLE{niu_lcss_2023_consensus_output_prediction,
  author  = {Niu, Kaicheng and Wardi, Yorai and Abdallah, Chaouki T. and Hayajneh, Muhammad},
  title   = {Consensus Controller for Heterogeneous Multi-Agent Systems Using Output Prediction},
  journal = {IEEE Control Systems Letters},
  year    = {2023},
  volume  = {7},
  pages   = {673--678},
  doi     = {10.1109/LCSYS.2022.3217160}
}

@INPROCEEDINGS{shivam_cdc2019_pursuit_evasion,
  author    = {Shivam, Shashwat and Kanellopoulos, Aris and Vamvoudakis, Kyriakos G. and Wardi, Yorai},
  booktitle = {2019 IEEE 58th Conference on Decision and Control (CDC)},
  title     = {A Predictive Deep Learning Approach to Output Regulation: The Case of Collaborative Pursuit Evasion},
  year      = {2019},
  pages     = {853--859},
  doi       = {10.1109/CDC40024.2019.9028950}
}

@INPROCEEDINGS{notomista_acc2024_safe_tracking,
  author    = {Notomista, Gennaro and Wardi, Yorai},
  booktitle = {2024 American Control Conference (ACC)},
  title     = {A Safe and Computationally Efficient Tracking Control Algorithm for Autonomous Vehicles},
  year      = {2024},
  pages     = {3734--3739},
  doi       = {10.23919/ACC60939.2024.10644288}
}

@inproceedings{niu_cdc2025_stability_flat_arxiv,
  author    = {Niu, Kaicheng and Wardi, Yorai and Abdallah, Chaouki T.},
  title     = {Stability Analysis of the Newton--Raphson Controller for a Class of Differentially Flat Systems},
  booktitle = {Proc.\ IEEE Conference on Decision and Control (CDC)},
  year      = {2025},
}

@software{jax2018github,
  author = {James Bradbury and Roy Frostig and Peter Hawkins and Matthew James Johnson and Chris Leary and Dougal Maclaurin and George Necula and Adam Paszke and Jake Vander{P}las and Skye Wanderman-{M}ilne and Qiao Zhang},
  title = {{JAX}: composable transformations of {P}ython+{N}um{P}y programs},
  url = {http://github.com/jax-ml/jax},
  version = {0.3.13},
  year = {2018},
}

@article{behnel2011cython,
  title={Cython: The best of both worlds},
  author={Behnel, Stefan and Bradshaw, Robert and Citro, Craig and Dalcin, Lisandro and Seljebotn, Dag Sverre and Smith, Kurt},
  journal={Computing in Science \& Engineering},
  volume={13},
  number={2},
  pages={31--39},
  year={2011},
  publisher={IEEE}
}

@article{QT-JW-ZX-TL:2021,
  title={Swing-reducing flight control system for an underactuated indoor miniature autonomous blimp},
  author={Tao, Q. and Wang, J. and Xu, Z. and Lin, T. X. and Yuan, Y. and Zhang, F.},
  journal={IEEE/ASME Transactions on Mechatronics},
  volume={26},
  number={4},
  pages={1895-1904},
  year={2021},
  publisher={IEEE}
}

@book{TF:2011,
  title={Handbook of marine craft hydrodynamics and motion control},
  author={T. I. Fossen},
  year={2011},
  publisher={John Wiley \& Sons}
}

@article{QT-JT-JC-YY-FZ:2021,
  title={Modeling and control of swing oscillation of underactuated indoor miniature autonomous blimps},
  author={Q. Tao and J.T. Tan and J. Cha and Y. Yuan and F. Zhang},
  journal={Unmanned Systems},
  volume={9},
  number={01},
  pages={73--86},
  year={2021},
  publisher={World Scientific}
}

@inproceedings{QT-MH-FZ:2020,
  title={Modeling and identification of coupled translational and rotational motion of underactuated indoor miniature autonomous blimps},
  author={Q. Tao and M. Hou and F. Zhang},
  booktitle={16th International Conference on Control, Automation, Robotics and Vision},
  pages={339-344},
  year={2020},
  organization={IEEE}
}

@inproceedings{HF-KK-YH-FM-KK-HA:2007,
  title={State-predictive control of an autonomous blimp in the presence of time delay and disturbance},
  author={H. Fukushima and K. Kon and Y. Hada and F. Matsuno and K. Kawabata and H. Asama},
  booktitle={2007 IEEE International Conference on Control Applications},
  pages={188-193},
  year={2007},
  organization={IEEE}
}

@article{MW-AA-MC-IS-FS:2024,
  title={Robust Design of Sliding Mode Control for Airship Trajectory Tracking with Uncertainty and Disturbance Estimation},
  author={M. Wasim and A. Ali and M. A. Choudhry and I. Shaikh and F. Saleem},
  journal={Journal of Systems Engineering and Electronics},
  volume={35},
  number={1},
  pages={242-258},
  year={2024},
  publisher={BIAI}
}

@article{MK-LB-SC:2024,
  title={Feedback Linearization of an Underactuated Miniature Blimp With Zero Dynamics Mitigation Using High Order Control Barrier Functions},
  author={M. Kasmalkar and L. Baird and S. Coogan},
  journal={IEEE Control Systems Letters},
  year={2024},
  publisher={IEEE}
}

@inproceedings{TL-EP-MB-AA:2022,
  title={Deep residual reinforcement learning based autonomous blimp control},
  author={Liu, Yu Tang and Price, Eric and Black, Michael J and Ahmad, Aamir},
  booktitle={2022 IEEE/RSJ International Conference on Intelligent Robots and Systems (IROS)},
  pages={12566--12573},
  year={2022},
  organization={IEEE}
}

\end{document}